\documentclass[10pt,twocolumn,letterpaper]{article}
\usepackage{authblk}

\usepackage[pagenumbers]{wacv} 

\usepackage{graphicx}
\usepackage{amsmath}
\usepackage{amssymb}
\usepackage{booktabs}
\usepackage{amsfonts, amsmath, multicol, multirow, subcaption, booktabs, graphicx}
\usepackage[table]{xcolor}

\newcommand{\cmark}{\ding{51}}
\newcommand{\xmark}{\ding{55}}
\usepackage{pifont}

\usepackage[pagebackref,breaklinks,colorlinks]{hyperref}

\usepackage[capitalize]{cleveref}
\crefname{section}{Sec.}{Secs.}
\Crefname{section}{Section}{Sections}
\Crefname{table}{Table}{Tables}
\crefname{table}{Tab.}{Tabs.}

\usepackage[normalem]{ulem}

\def\wacvPaperID{533} 
\def\confName{WACV}
\def\confYear{2025}

\begin{document}
\date{}
\title{Vision-Aware Text Features in Referring Image Segmentation: \\ From Object Understanding to Context Understanding}

\author{Hai Nguyen-Truong\textsuperscript{1} \qquad E-Ro Nguyen\textsuperscript{2,3,4}\thanks{Co-first author} \qquad 
Tuan-Anh Vu\textsuperscript{1}\thanks{Corresponding author} \\  Minh-Triet Tran\textsuperscript{3,4} \qquad Binh-Son Hua\textsuperscript{5} \qquad Sai-Kit Yeung\textsuperscript{1}}


\affil{
\textsuperscript{1}The Hong Kong University of Science and Technology \quad
\textsuperscript{2}Stony Brook University \\
\textsuperscript{3}University of Science, VNU-HCM, Ho Chi Minh City \\ 
\textsuperscript{4}Vietnam National University, Ho Chi Minh City \\ 
\textsuperscript{5}Trinity College Dublin
}

\maketitle


\begin{abstract} 
Referring image segmentation is a challenging task that involves generating pixel-wise segmentation masks based on natural language descriptions. The complexity of this task increases with the intricacy of the sentences provided. 
Existing methods have relied mostly on visual features to generate the segmentation masks while treating text features as supporting components. 
However, this under-utilization of text understanding limits the model's capability to fully comprehend the given expressions. 
In this work, we propose a novel framework that specifically emphasizes object and context comprehension inspired by human cognitive processes through Vision-Aware Text Features.
Firstly, we introduce a CLIP Prior module to localize the main object of interest and embed the object heatmap into the query initialization process.
Secondly, we propose a combination of two components: Contextual Multimodal Decoder and Meaning Consistency Constraint, to further enhance the coherent and consistent interpretation of language cues with the contextual understanding obtained from the image. 
Our method achieves significant performance improvements on three benchmark datasets RefCOCO, RefCOCO+ and G-Ref. Project page: \url{https://vatex.hkustvgd.com/}.
\end{abstract}

\section{Introduction}

Referring image segmentation (RIS) is an emerging new task in computer vision that predicts pixel-wise segmentation of visual objects in images from natural language cues.  
Compared to traditional segmentation ~\cite{segmentation1, segmentation2, segmentation3, vis1, vis2, vis3, vis5}, RIS allows users to select and control the segmentation results via text prompts, which is useful in various applications such as image editing, where users can modify specific parts of an image using simple text commands, and in robotics, where robots need to understand and act on verbal instructions in dynamic environments.

A particular technique to solve RIS is to obtain a robust alignment between language and vision. Performing such an alignment presents significant challenges due to the nature of languages, which are highly ambiguous without given context. Early alignment approaches~\cite{early1, early2, early3, early4} in RIS either used bottom-up methods, merging vision and language features in early fusion and using an FCN~\cite{fcn} as a decoder to produce object masks, or top-down methods, which first identify objects in image and use the expression as the grounding criterion to select best-matched result.

\begin{figure}[!t]
    \centering
    \includegraphics[width=\linewidth]{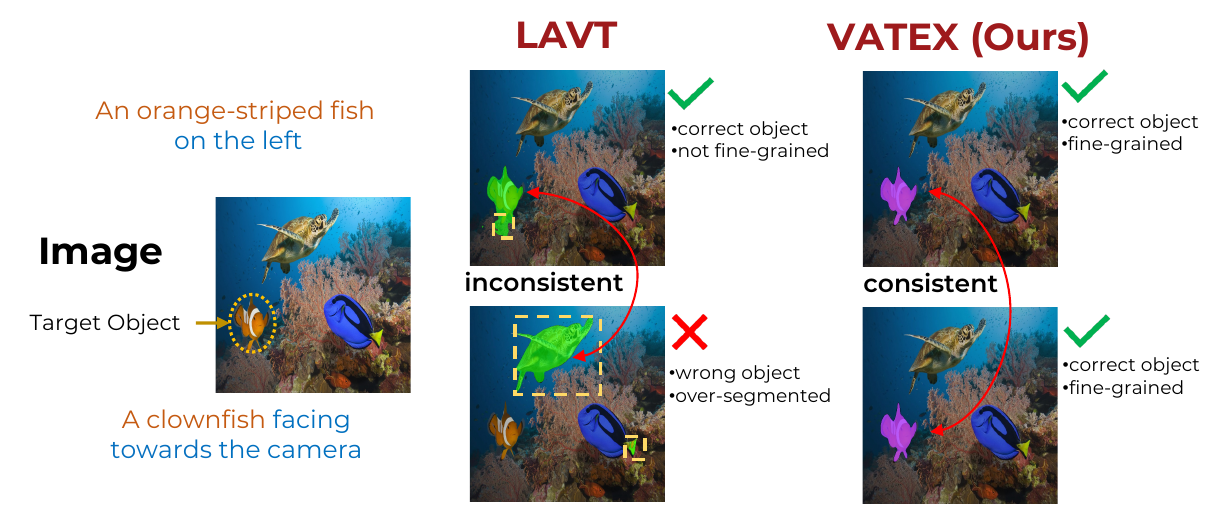}
    \caption{Qualitative comparison between LAVT and Ours. The yellow box indicates the wrong segmentation results. \textcolor[HTML]{C55A11}{Object understanding} and \textcolor[HTML]{0570C0}{Context understanding} are required to tackle the challenge of complex and ambiguous language expression.}
    \label{fig:teaser}
\end{figure}

Recent approaches~\cite{lavt, vlt_tpami, polyformer} are based on transformers that learn the interaction between vision-text modalities followed by a standard encoder-decoder process to produce pixel-level segmentation results. However, existing methods have relied mostly on visual features to generate the segmentation masks while treating text features as supporting components in the fusion module. This insufficient utilization of text understanding hampers these methods' ability to accurately segment target objects for challenging expressions involving rare object vocabulary or contextual relationships between objects. For instance, as illustrated in ~\Cref{fig:teaser}, while LAVT~\cite{lavt} can effectively segment the well-defined "orange-striped fish" and identify specific location information "on the left" in the first expression, it struggles with the expression "clownfish" and incorrectly identifies the turtle that is "facing towards the camera" in the second expression. This inconsistency highlights the limitations of current approaches in understanding complex expressions, especially in handling unseen vocabulary and varying expressions referring to the same object.

Moreover, the human approach to RIS~\cite{cog_science1, cog_science2, cog_science3} does not involve parsing or understanding complex sentences entirely. Instead, we naturally break down a referring expression into its core components: the object of interest and its description with context information. Initially, the primary focus is on identifying what is the object mentioned in the expression (e.g., the main object of interest). Following this, the search space within the image is narrowed to objects that match the main object's category. The final step involves using the specific characteristics or contextual information described in the expression to pinpoint the target object. Inspired by this, we propose decomposing this task into two processes: \textcolor[HTML]{C55A11}{object understanding} and \textcolor[HTML]{0570C0}{context understanding}. This decomposition allows for a more comprehensive understanding of text features, ultimately enhancing the accuracy and consistency of referring expression segmentation.

Firstly, in terms of \textit{object understanding}, current methods do not utilize the object representation in the query initialization process. ReferFormer~\cite{referformer} generated object queries conditioned on language expressions, while VLT~\cite{vlt_iccv} implicitly employed multiple query vectors with different attention weights to generate various interpretations of the language description. However, these variations may lead to confusion and conflict with each other and may not focus on the target object. On the other hand, we propose CLIP Prior to explicitly integrate visual information of the primary object of interest into text cues during the query initialization process. This module transfers the knowledge from pre-trained model CLIP~\cite{clip} and generates an object-centric visual heatmap to create adaptive, vision-aware queries, enhancing generalization and robustness of object comprehension, even in the challenging case where the expression contains "unseen" category (\eg clownfish).

Secondly, for \textit{context understanding}, we introduce a Contextual Multimodal Decoder (CMD) to further exploit the superior interaction between visual and text modalities, especially the vision-to-language interaction. CMD aims to enhance text features by using contextual information obtained from the visual features and to bring the semantic-aware textual information back to visual features in a hierarchical architecture. While we can use the ground truth mask annotations as a direct learning signal to supervise the language-to-vision features, the opposite interaction is implicitly learned without any learning signal. By observing that there are multiple ways to describe an instance based on the context provided by the image, we propose the Meaning Consistency Constraint (MCC) as a contrastive learning signal to enforce the consistency of vision-aware text features produced from CMD among different expressions referring to the same instance in an image. The vision-to-language interaction can explicitly learn through this extra in-context learning signal, resulting in a \textit{profound, coherent, and contextual understanding} in the feature space.

Our method is evaluated on three widely-used image datasets, RefCOCO, RefCOCO+, and G-Ref, and further extends the results to video datasets, Ref-Youtube-VOS and Ref-DAVIS17. These datasets consist of diverse and challenging text expressions, and our proposed model achieves state-of-the-art performance on all five. Through various ablation studies, we have demonstrated the effectiveness of our model and shown that it can achieve robust referring segmentation even in challenging scenarios. 

Our main contributions can be summarized as follows:
\begin{itemize}
    \item We address the current limitations of existing methods in dealing with complex text expressions and present a novel framework to utilize \textbf{V}ision-\textbf{A}ware \textbf{TEX}t Features (\textbf{VATEX}) for a better understanding of text expressions in RIS by decomposing it into Object Understanding and Context Understanding components.
    \item We introduce a novel CLIP Prior to embed an object-centric visual heatmap in the query initialization process, enhancing object understanding by transferring knowledge from the pre-trained CLIP model.
    \item We propose Contextual Multimodal Decoder (CMD) followed by a Meaning Consistency Constraint (MCC) as a learning signal for vision-to-language branch to improve context understanding. CMD module enhances the interaction between visual and text modalities, while MCC ensures consistent interpretation of different expressions conditioned in an image.
    \item Our method achieves superior performance on all splits of the RefCOCO, RefCOCO+, and G-Ref for image datasets and Ref-YouTube-VOS and Ref-DAVIS 2017 for video datasets, surpassing the current state of the art for each dataset, especially in datasets with the more complex expressions.
\end{itemize}

\begin{figure*}[!t]
    \centering
    \includegraphics[width=0.99\linewidth]{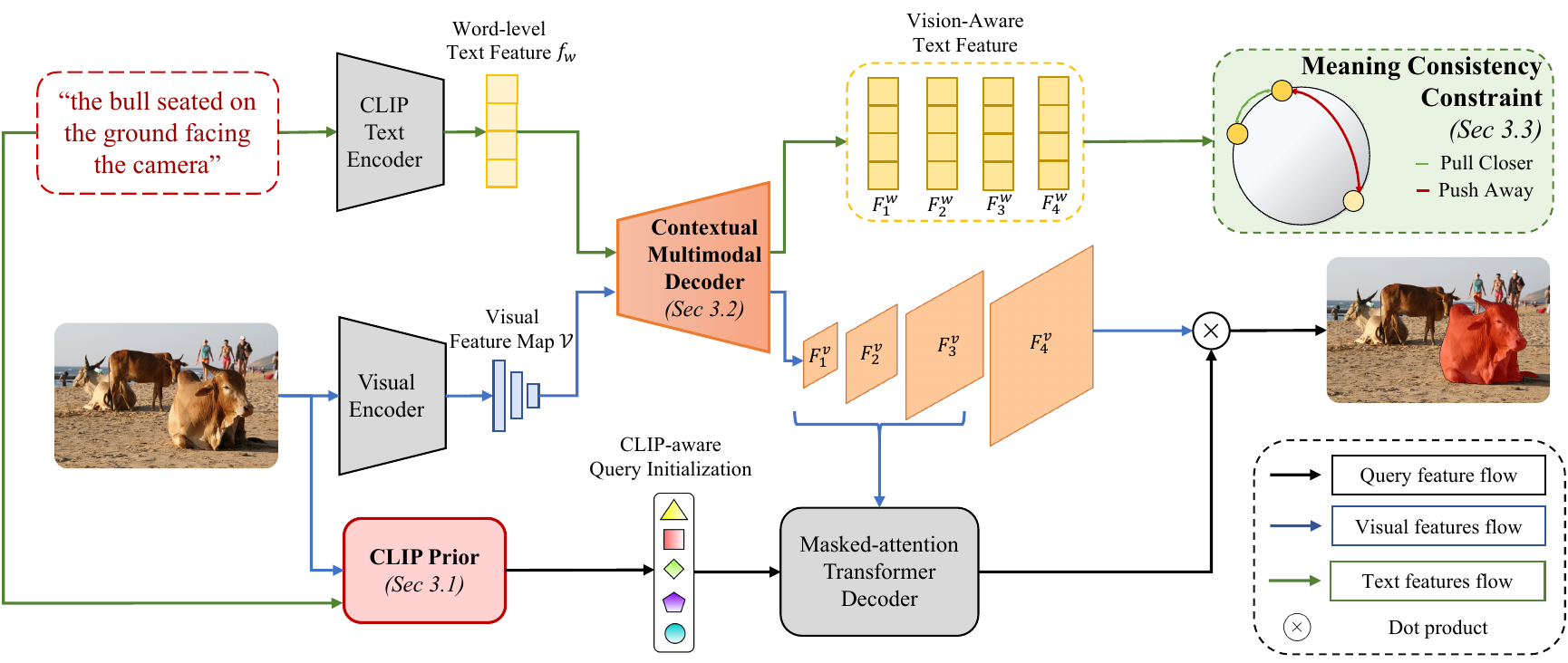}
    \caption{The overall architecture of VATEX, which processes input images and language expressions through two concurrent pathways. Initially, the CLIP Prior module generates object queries, while simultaneously, traditional Visual and Text Encoders create multiscale visual feature maps and word-level text features. These visual and text features are passed into the Contextual Multimodal Decoder to enable multimodal interactions, yielding vision-aware text features and text-enhanced visual features. We then harness vision-aware text features to ensure semantic consistency across varied textual descriptions that reference the same object by employing sentence-level contrastive learning, as described in the Meaning Consistency Constraint section. On the other hand, the text-enhanced visual features and the object queries generated by the CLIP Prior are refined through a Masked-attention Transformer Decoder to produce the final output segmentation masks.}
    \label{fig:overview}
\end{figure*}

\section{Related Work}
\label{sec:related_works}

\noindent\textbf{Referring image segmentation}~\cite{hu2016segmentation} aims to generate pixel-wise segmentation masks for referred objects in images given a natural language expression. 
Early works~\cite{liu_learning_2019, huang_referring_2020, feng_encoder_2021} proposed to extract visual and linguistic features independently from convolutional and recurrent neural networks, respectively, and then concatenating these features to create multimodal features for decoding final segmentation results. 
In recent works~\cite{reftr, seqtr, lavt, vlt_iccv, lbdt, mttr, xdecoder, hipie, etris, coupalign}, transformer-based multimodal encoders have been designed to fuse visual and text features, capturing the interaction between vision and language information in the early stage. VG-LAW~\cite{vglaw} utilizes language-adaptive weights to dynamically adjust the visual backbone, enabling expression-specific feature extraction for better mask prediction, while LISA~\cite{lisa} learns to generate segmentation masks based on Large Language Models. PolyFormer~\cite{polyformer}, on the other hand, treats this task as a sequential polygon generation. Another line of works have explored enhancing text understanding in RIS using graph-based methods: CMPC-RefSeg~\cite{cmpc-refseg} classifies words into entity, attribute, relation, and other categories, building a graph with entities and attributes as nodes and relations as edges, while LSCM-RefSeg~\cite{lscm-refseg} constructs fully connected graph, then based on dependency parsing trees to prune unnecessary edges. In contrast, our model does not depend on graph convolutional networks, instead focusing on utilizing the vision-aware text features in text understanding.

\noindent\textbf{Query Initialization.} The DETR (DEtection TRansformer) framework~\cite{detr} has achieved impressive performance in object detection by directly transforming the task of object detection into a set prediction problem. Building upon DETR, several works have focused on improving the query initialization process for better performance. Deformable DETR~\cite{deformable-detr} proposes a deformable transformer architecture to refine object queries, while DAB-DETR~\cite{dab-detr} directly uses the bounding box coordinate in the image to improve query initialization. In the field of referring segmentation, ReferFormer~\cite{referformer} extracts the word embeddings from the referring expression and treats them as the initial query for the framework. Our CLIP Prior elevates this approach by incorporating a CLIP-generated heatmap, enriching the textual features with spatial context during query initialization. This enriched query leads to improved performance in the subsequent segmentation stages.

\noindent\textbf{Contrastive Learning} is pivotal in advancing vision-language tasks~\cite{cl1, cl2, cl3, cl4}, enhancing model performance by distinguishing similarities and differences in visual and textual data. CLIP~\cite{clip} employed a contrastive loss on an extensive image-text dataset. CRIS~\cite{cris} leveraged text and pixel-level contrastive learning while VLT~\cite{vlt_tpami} applied masked contrastive learning to refine visual features across diverse expressions. Unlike previous approaches~\cite{cris, vlt_tpami} that solely focus on improving visual qualities by raw linguistic information, our work utilizes contrastive learning to enhance the comprehension of varied expressions conditioned in a shared image context before using it to enrich the visual features. This ensures the accuracy and stability of mutual interaction between text and visual features, particularly through the comparison of vision-aware expressions related to objects in an image.

\section{Proposed Method}
\label{sec:proposed_method}

Inspired by the human approach to RIS, which involves breaking down a referring expression into its core components: object of interest and contextual description, we propose simplifying the text expression by decomposing it into object and context parts. This decomposition aims to enhance the text understanding, thus improving the accuracy and consistency of referring expression segmentation. 

Our framework is constructed by three main components, as demonstrated in~\Cref{fig:overview}. 
First, for object understanding, we propose a CLIP Prior module to generate an object-centric visual heatmap that localizes the object of interest from the text expression, which can be subsequently used to initialize the object queries for the DETR-based method (\Cref{sec:clip_prior}). 
Next, we utilize cross-attention modules to interact between visual-text modalities in a hierarchical architecture via our Contextual Multimodal Decoder (\Cref{sec_cmd}) and leverage Meaning Consistency Constraint to harness vision-aware text features generated by CMD (\Cref{sec_mcc}). 
We further adopt a masked-attention transformer decoder~\cite{mask2former} to enhance the object queries through multiscale text-guided visual features. 
Finally, the enhanced object queries and the visual features from CMD are utilized to output segmentation masks (\Cref{sec:instance_matching}).

Mathematically, given an input image with the size of $H \times W \times 3$, we can obtain the multiscale visual feature maps $\mathcal{V} = \left\{V_i\right\}_{i = 1}^4, V_i \in \mathbb{R}^{H_i \times W_i \times C_i}$ from the Visual Encoder that captures the visual information in the data, where $H_i, W_i, C_i$ denote the height, width, and the channel dimension of $V_i$. Given the $L$-word language expression as input, we use our Text Encoder to encode it into word-level text features $f_w \in \mathbb{R}^{L \times C} $ with $C$ as the channel dimension. 
Our visual and text features will be further processed as described in the following sections.

\subsection{Object Localization with CLIP Prior}
\label{sec:clip_prior}

\begin{figure}[!t]
    \centering
    \includegraphics[width = \linewidth]{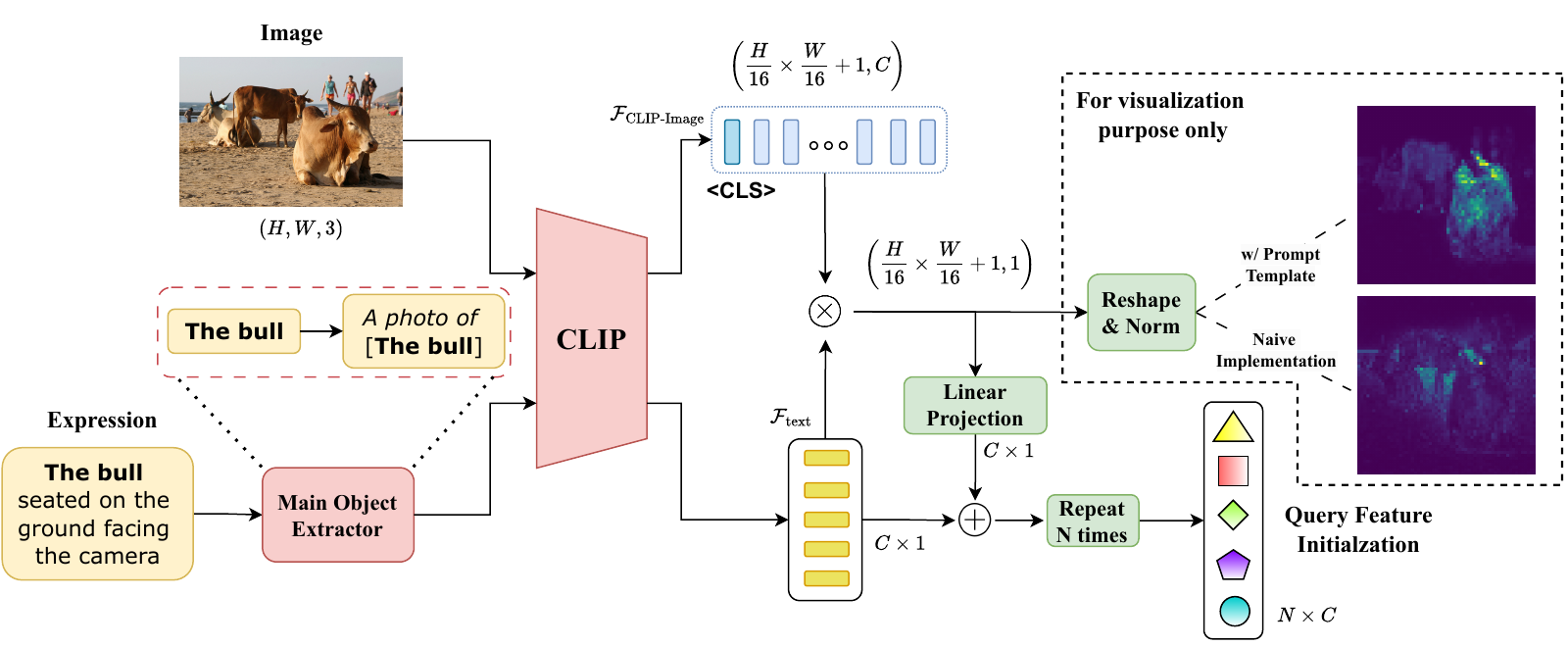}
    \caption{Our CLIP Prior exploits the alignment of CLIP-Image and CLIP-Text embeddings for better query initialization. Best viewed with zoom.}
    \label{fig:CLIPPrior}
\end{figure}

For the object part, we first extract the main noun phrase from the expression using spaCy~\cite{spacy2} (e.g., the bull) in order to focus only on this main noun phrase. Referring expressions in the RIS task are object-centric, which means that the main noun phrase is the main object of the sentence. We then convert the complex referring expression to a simple template-based sentence before passing it to the CLIP Encoder. In our implementation, we use "A Photo of [Object]" as our template as it is the most common prompt to describe an object~\cite{Zhou2022ijcv} in CLIP, where the resulting text feature is represented by $\mathcal{F}_{\text{text}}$. We found that this improves the accuracy of the heatmap in localizing the object of interest. In a separate flow, our input image goes through the CLIP Visual Encoder, resulting in features for multiple image tokens $\mathcal{F}_{\text{CLIP-Image}} \in \mathbb{R}^{\left(\frac{H}{16} \times \frac{W}{16} + 1\right) \times C}$.

Our visual heatmap for the object of interest can be obtained by calculating the similarity between the visual and text features, then reshaping to image space and going through L2-normalization:
\begin{equation}
\text{Heatmap = norm} \left(\frac{\mathcal{F}_{\text{CLIP-image}}}{\Vert \mathcal{F}_{\text{CLIP-image}} \Vert} \cdot \frac{\mathcal{F}_{\text{text}}}{\Vert \mathcal{F}_{\text{text}}\Vert}\right).
\end{equation}

As normal practice, positional prior from CLIP is embeded to the text features by changing the dimension of the similarity map from $\frac{H}{16} \times \frac{W}{16} + 1$ to $C$, then repeat it $N$ times together with the text features, then add these two to create initial object queries feature with $N$ queries, each with $C$-dimension, for embedding the positional prior and text information into the query feature.  Unlike previous methods that typically learn the target object representation \textit{implicitly} through multi-modal transformers~\cite{vlt_iccv} or convert only \textit{textual information} from natural language into object queries~\cite{referformer}, our approach explicitly generates the heatmap and embeds it in the query initialization process. This ensures that the queries contain rich and useful information about both visual and textual aspects, as well as the alignment between these modalities. Such a comprehensive query initialization is essential for effective object understanding, particularly in challenging scenarios involving complex or unseen vocabulary.

While this template approach can efficiently localize regions of interest, it may lead to information loss due to oversimplification (e.g., focusing only on the bull in this scenario). However, its primary function is to narrow down the search space by \textit{localizing a region} containing the object of interest, not necessarily finding out the exact object. To find the exact object of interest, the full-text expression needs to pass through the CMD and MCC for more comprehensive characteristics and contextual understanding.

\subsection{Contextual Multimodal Decoder} 
\label{sec_cmd}

Contextual Multimodal Decoder (CMD) is proposed to produce multi-scale text-guided visual feature maps while enhancing contextual information from the image into word-level text features in a hierarchical design, see the architecture figure in the supplementary material. 
Specifically, our CMD is based on a feature pyramid network architecture~\cite{fpn}, which has four levels. Each level transfers the semantic information from visual features to text features and then uses these vision-aware text features to update the visual features afterward via cross-modal attention. 

In the $i$-th level of CMD, given the input visual features $V_i$ and text features $F^w_{i - 1}$, the multi-modal interactions are performed in two steps. First, a cross-attention that takes text features  $F^w_{i - 1}$ as query and visual features $V_i$ as key and value is used to model the relationship of the text and visual information. Then it forms the vision-aware text features by associating them with current text features: 
\begin{equation}
    F^w_i = \text{MHA}(F^w_{i - 1}, V_i, V_i) \cdot F^w_{i - 1},
\end{equation}
where $\text{MHA}(q, k, v)$ is the multi-head cross-attention module with query $q$, key $k$, value $v$.
 
$F^w_i$ is then treated as the key and value and $V_i$ is treated as the query in another multi-head cross-attention module to reinforce the alignment between the visual and text modalities and generate features $V'_i$. Consequently, $V'_i$ is fused with the text-guided visual feature $F^v_{i - 1}$ from the previous level $i - 1$ followed by a Conv2d layer to obtain the text-guided visual feature $F^v_i$. Mathematically, the whole process is as follows:
\begin{align}
    V'_i &= \text{MHA}(V_i, F^w_i, F^w_i) \cdot V_i,\\ 
    F^v_i &= \text{Conv2d}(V'_i + \text{Ups}(F^v_{i - 1})),
\end{align}
where $\text{Conv2d}()$ is the 2D convolutional layer, and $\text{Ups}()$ denotes upsampling $F^v_{i - 1}$ to the size of $V'_i$.
For the first level with $i = 1$, we skip $F^v_0$ and let $F^w_0 = f_w$, where $f_w$ is the word-level linguistic features extracted by the Text Encoder. 

Previous methods have developed various bidirectional multimodal fusion mechanisms, including word-pixel alignment in encoder stage~\cite{etris, coupalign} and region-language interactions~\cite{rela}. Compared to these approaches, our novelty is the combination of CMD and MCC, where MCC serves as an in-context learning signal to enrich the vision-aware text features and further enhance the vision-language interactions within the hierarchical architecture of CMD.

\begin{table*}[!t]
\begin{minipage}{.62\linewidth}
\centering
\caption{{Quantitative results of referring image segmentation on Ref-COCO, Ref-COCO+, G-Ref datasets on mIoU metric.}}
\setlength{\tabcolsep}{3pt}
\renewcommand{\arraystretch}{1.3}
\resizebox{\linewidth}{!}{%
\begin{tabular}{l|c|c|ccc|ccc|cc}
\toprule

\multirow{2}{*}{\textbf{Method}} & \multicolumn{2}{c|}{\textbf{Backbone}} & \multicolumn{3}{c|}{\textbf{RefCOCO}} & \multicolumn{3}{c|}{\textbf{RefCOCO+}} & \multicolumn{2}{c}{\textbf{G-Ref}} \\ \cline{2-11} 
                                    &     Visual      &    Textual             & val      & testA   & testB   & val      & testA    & testB   & val         & test        \\ \midrule
CRIS~\cite{cris}                & ResNet-101  & CLIP & {70.47} & {73.18}  & 66.10  & {62.27} & {68.08}  & 53.68  & {59.87} & {60.36}     \\ 
CM-MaskSD~\cite{cmmasksd} & CLIP-ViT-B & CLIP & 72.18 & 75.21 & 67.91 & 64.47 & 69.29 & 56.55 & 62.67 & 62.69 \\ 
VLT~\cite{vlt_tpami} & Swin-B & BERT & 72.96 & 75.96 & 69.60 & 63.53 & 68.43 & 56.92 & 63.49 & 66.22 \\ 
LAVT~\cite{lavt}                & Swin-B    & BERT & 74.46 &	76.89 &	70.94 &	65.81 &	70.97	& 59.23 & 63.34 &	63.62 \\ 
LISA-7B~\cite{lisa} & ViT-H SAM & Vicuna-7B &  74.10 & 76.50 & 71.10 & 65.10 & 67.40 & 56.50 & 66.40 & 68.50 \\
VG-LAW~\cite{vglaw} & ViT-B & BERT & 75.05  & 77.36 & 71.69 & 66.61 & 70.30 & 58.14 & 65.36 & 65.13 \\ 
\midrule
\rowcolor{gray!15} \textbf{VATEX (Ours)}                   & Swin-B  & CLIP & \textbf{78.16} & \textbf{79.64}  & \textbf{75.64}  & \textbf{70.02} & \textbf{74.41}  & \textbf{62.52}  & \textbf{69.73} & \textbf{70.58} \\  

\bottomrule
\end{tabular}
}
\label{tab:ref-3datasets}
\end{minipage}
\begin{minipage}{.01\linewidth}
    \textbf{}
\end{minipage}
\begin{minipage}{.36\linewidth}
        \centering
        \caption{\footnotesize{Precision analysis at different threshold value comparison between VATEX and recent SOTA methods.}}
        \vspace{-2.5mm}
        \resizebox{0.9\linewidth}{!}{%
        \setlength{\tabcolsep}{4pt}
        \renewcommand{\arraystretch}{1.2}
            \begin{tabular}{l|ccccc|c}
                    \toprule
                    \textbf{Methods} & Pr@0.5 & Pr@0.7 & Pr@0.9 & mIoU \\ 
                    \midrule
                    LAVT~\cite{lavt} & 84.46  & 75.28  & 34.30 & 74.46 \\ 
                    ReLA~\cite{rela} & 85.92  & 77.71  & 34.99 & 75.61 \\ 
                    CG-Former~\cite{cgformer} & 87.23   & 78.69  & 38.77 & 76.93 \\ 
                   \rowcolor{gray!15} \textbf{VATEX (Ours)} & \textbf{88.12}  & \textbf{82.54}  & \textbf{45.11} & \textbf{78.17} \\ 
                    \bottomrule
            \end{tabular}%
        }
		\vspace{4mm}
		\label{tab:precision-analysis} 
		\caption{\footnotesize{{Quantitative results on video datasets. }}}
		\vspace{-2mm}
		\resizebox{\linewidth}{!}{%
		\setlength{\tabcolsep}{4pt}
		\renewcommand{\arraystretch}{1.2}
			\begin{tabular}{l|ccc|ccc}
				\toprule 
				\multirow{2}{*}{\textbf{Methods}} & \multicolumn{3}{c|}{\textbf{Ref-YT-VOS}} & \multicolumn{3}{c}{\textbf{Ref-DAVIS17}}                                           \\ \cline{2-7} 
				 & \multicolumn{1}{c|}{$\mathcal{J}\&\mathcal{F}$} & \multicolumn{1}{c|}{$\mathcal{J}$} & \textbf{$\mathcal{F}$}  &  \multicolumn{1}{c|}{$\mathcal{J}\&\mathcal{F}$} & \multicolumn{1}{c|}{$\mathcal{J}$} & \textbf{$\mathcal{F}$} \\ \midrule
				ReferFormer~\cite{referformer} & \multicolumn{1}{c|}{62.9}            & \multicolumn{1}{c|}{61.3}       & 67.5              & \multicolumn{1}{c|}{61.1}            & \multicolumn{1}{c|}{58.1}       & 64.1  \\
				VLT ~\cite{vlt_tpami} & \multicolumn{1}{c|}{63.8}            & \multicolumn{1}{c|}{61.9}       & 65.6             & \multicolumn{1}{c|}{61.6}            & \multicolumn{1}{c|}{58.9}       & 64.3       \\
			   \rowcolor{gray!15} \textbf{VATEX (Ours)} & \multicolumn{1}{c|}{\textbf{65.4}}            & \multicolumn{1}{c|}{\textbf{63.3}}       & \textbf{67.5}     & \multicolumn{1}{c|}{\textbf{65.4}}   & \multicolumn{1}{c|}{\textbf{62.3}}  & \textbf{68.5}  \\ 
			\bottomrule
			\end{tabular}
		}
		\label{tab:video-results}          
\end{minipage}
\end{table*}

\subsection{Meaning Consistency Constraint} 
\label{sec_mcc}

\begin{figure}[!t]
    \centering
    \includegraphics[width=\linewidth]{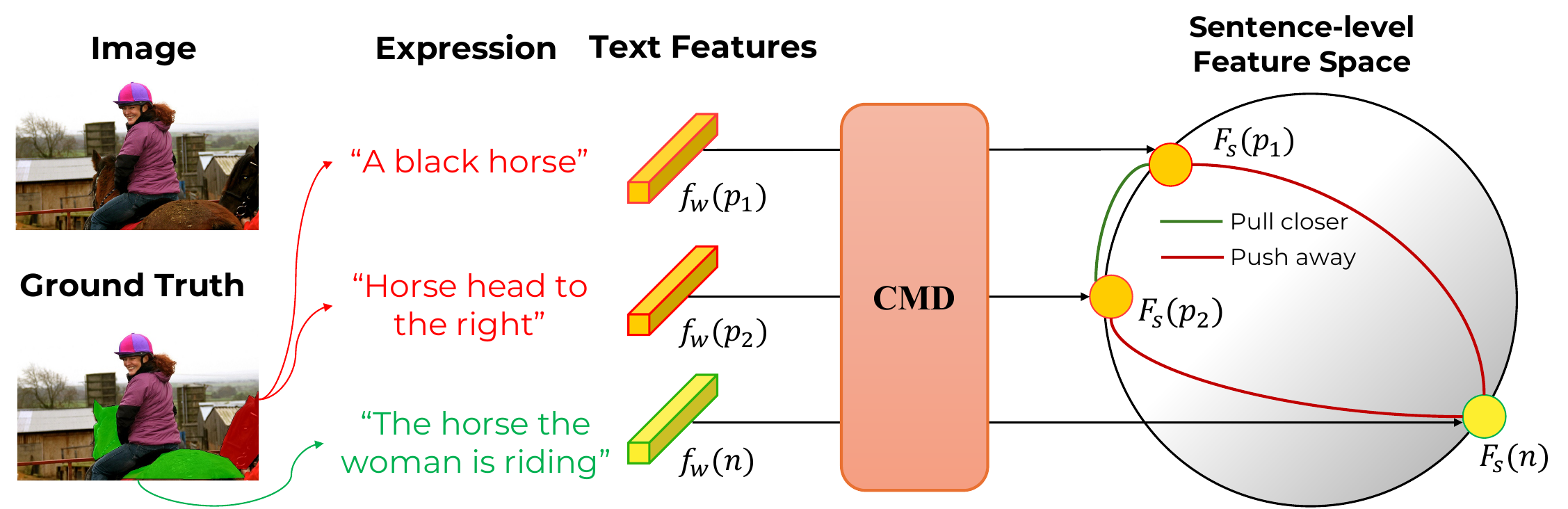}
    \caption{Illustration of Meaning Consistency Constraint. Vision-aware text embeddings of different expressions are passed through a contrastive learning module in sentence-level feature space. Embeddings referring to the same object are pulled closer while pushing others far away. Best view in color.}
    \label{fig:meaning_consistency}
\end{figure}

Each object in an image can be described by various text expressions. Although the linguistic meaning of these descriptions may be different, they should convey the same semantic meaning when referencing that image (see two red expressions in ~\Cref{fig:meaning_consistency}). According to this perspective, it's crucial for CMD to gradually comprehend contextual cues from visual features into textual representations and ensure consistent identification of target objects, where expressions referring to the same object yield identical representations. However, previous studies have often overlooked the relationship between expressions that pertain to the same instance.

To delve deeper into this relationship and provide the explicit in-context learning signal for vision-aware text features within CMD, we propose Meaning Consistency Constraint (MCC), a sentence-level contrastive learning approach. MCC aims to learn meaningful and discriminative representations for different expressions while consistently pulling sentences referring to the same object close to each other. 

Unlike previous contrastive learning-based approaches~\cite{vlt_tpami, cris}, we focus on linguistic features that are enriched and conditioned by visual information. This can encourage CMD module to gradually learn how to produce richer text features and lead to the improvement of visual features in context understanding due to the bidirectional attention mechanism and hierarchical design of CMD module. 

Our contrastive learning pipeline is illustrated in ~\Cref{fig:meaning_consistency}. During training, we construct a triplet of text expressions for each image. Each triplet comprises two sentences that refer to the same object (positive samples), along with a third sentence describing a different object (negative sample). We denote the positive samples by $p_1, p_2$ and the negative sample by $n$, respectively. With each sample $x$, we derive the sentence-level feature by averaging the vision-aware word-level textual features:
\begin{equation}
    F_s(x) = \text{Avg}(F^w_4(x), \text{dim} = 0),
\end{equation}
where $F^w_4 \in R^{L \times C}, F_s \in R^{C}$.
We adopt the InfoNCE loss~\cite{infonce} to ensure that linguistic features referring to the same object converge, while features of different objects diverge:
\begin{equation}
    \mathcal{L}_{mcc} = -\log\left(\frac{\text{sim}(p_1, p_2)}{\text{sim}(p_1, n) + \text{sim}(p_2, n)} \right),
\end{equation}
where $\text{sim}(p, n) = \text{exp}(F_s(p) \cdot F_s(n))$ to calculate the exponential for cosine similarity of sentence-level obtained from the text expressions.

\subsection{Network Training}
\label{sec:loss}

\noindent\textbf{Prediction Heads.}
We adopt the masked-attention transformer decoder~\cite{mask2former} to update our initial query feature $f_o$ by using the multi-scale text-guided visual features $\left\{F^v_i\right\}_{i = 1}^3$ to obtain the final object queries $F_o \in \mathbb{R}^{N \times C}$. The final object queries will directly predict the probability of the target object $\hat{p} \in \mathbb{R}^{N}$. The high-resolution segmentation mask $\hat{s} \in R^{\frac{H}{4} \times \frac{W}{4} \times N}$ is produced by associating between object queries $F_o$ with the last fine-grained text-guided visual features $F^v_4 \in \mathbb{R}^{\frac{H}{4} \times \frac{W}{4} \times C}$, which can be formulated as:
\begin{equation}
    \hat{s} = \text{Sigmoid}(F^v_4 \cdot {F_o^\top}).
\end{equation}

\noindent\textbf{Instance Matching.}
\label{sec:instance_matching}
The prediction set output from prediction heads is represented by $\hat{y} = \left\{\hat{y_i}\right\}_{i = 1}^{N}$, where $\hat{y}_i = \left\{\hat{p_i}, \hat{s_i}\right\}$. 
Since a text expression refers to only a specific object, we denote the ground truth object as $y = \left\{p_{gt} = 1, s_{gt}\right\}$. The best prediction $\hat{y}_{\delta}$ can be found by a Hungarian algorithm~\cite{hungarian} by minimizing the matching cost in terms of probability and segmentation mask~\cite{maskformer, mask2former}. 

\vspace{2mm}
\noindent\textbf{Training.}
Our prediction $\hat{y}_{\delta}$ is supervised by three losses. Firstly, the class loss $\mathcal{L}_{cls}$ is binary cross entropy (BCE) loss to supervise the probability of the referred object. Secondly, the mask loss $\mathcal{L}_{mask}$ is a combination of dice loss and BCE. Finally, our sentence-level contrastive loss $\mathcal{L}_{mcc}$ is used to enforce our Meaning Consistency Constraint. The total loss can be formulated as follows:
\begin{equation}
    \mathcal{L}_{total} = \gamma_{cls}\mathcal{L}_{cls} + \gamma_{mask}\mathcal{L}_{mask} + \gamma_{mcc}\mathcal{L}_{mcc}, 
\end{equation}
where $\gamma_{cls}, \gamma_{mask}, \gamma_{mcc}$ are the scalar coefficients.

\vspace{2mm}
\noindent\textbf{Inference.} In inference, our method aligns with the standard practice of using a single image or video with one text expression, and MCC only requires sampling positive and negative expressions in the training phase. During inference, the query with the highest probability score is selected as the target object for the final output.

\section{Experimental Results}
\label{sec:Experiments}

\subsection{Experiment Setup}

We evaluate the performance of our model on three image datasets: RefCOCO~\cite{refcoco}, RefCOCO+~\cite{refcoco}, G-Ref~\cite{gref} and further evaluate the performance of our model on two video datasets: Ref-Youtube-VOS~\cite{urvos} and Ref-DAVIS17~\cite{davis}. For evaluation metrics, we follow previous work to use mean IoU (mIoU) and Precision at different thresholds (Pr@X) for image and $\mathcal{J}\&\mathcal{F}$ for video datasets.

During training, we freeze the CLIP model. Images are resized to a short side of 480. We set the coefficients for the losses as $\gamma_{cls} = 2, \gamma_{mask} = 5$, and $\gamma_{mcc} = 2$, with the feature dimension $C$ set to $256$. We train the network for $100,000$ iterations on the RefCOCO(+/g) datasets with an initial learning rate of $10^{-4}$ and is reduced by a factor of $0.1$ at the $2/3$ last iteration. For the Ref-Youtube-VOS dataset, we initialize the pre-trained weight from RefCOCO(+/g) and train the network for $100,000$ iterations. Regarding the Ref-DAVIS17 dataset, we directly use the weight obtained from the Ref-Youtube-VOS dataset for inference. The training process uses 2 NVIDIA RTX 3090 GPUs with a batch size of 24.
For detailed information on each dataset and implementation, please see the supplementary material.

 \begin{figure*}[!t]
    \centering
    \includegraphics[width=0.99\linewidth]{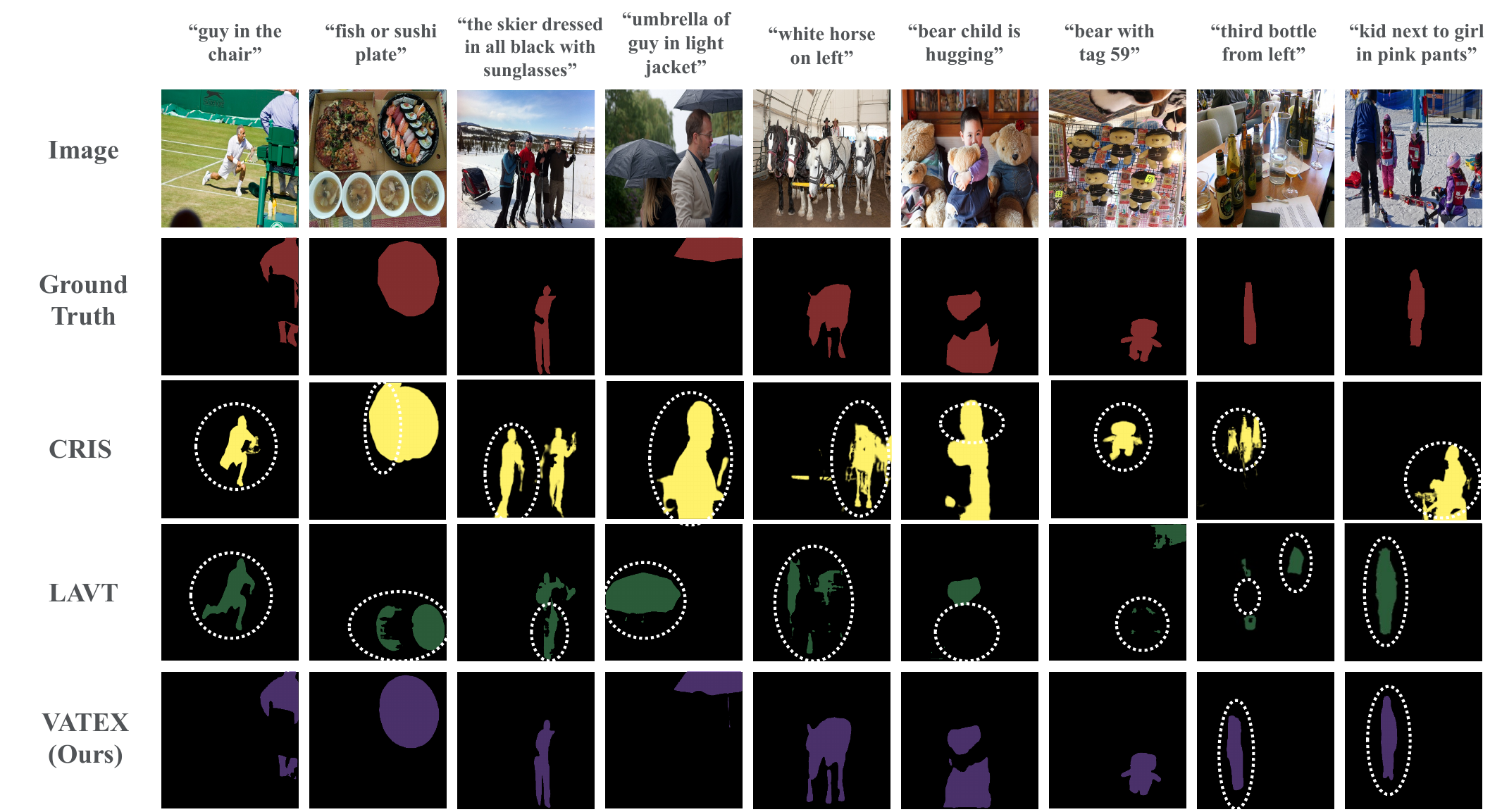}
    \caption{Results on RefCOCO(+/g) datasets. We compare our results with CRIS and LAVT. Our method excels at segmenting objects in complex scenarios, such as distinguishing between similar objects and localizing specific instances within a scene. The last two columns of the results show failure cases. Best viewed in color.}
    \label{fig:visualize}
\end{figure*}

\begin{table*}[t]

\begin{minipage}{.36\linewidth}
            \centering
            \footnotesize
            \caption{\footnotesize{Ablation Study on the val set of RefCOCO (mIoU) and Ref-YouTube-VOS ($\mathcal{J} \& \mathcal{F}$).}}
            \vspace{-2mm}
            \resizebox{\textwidth}{!}{%
            \setlength{\tabcolsep}{3pt}
            \renewcommand{\arraystretch}{1.15}
            \begin{tabular}{ccc|cc}
            \toprule
            \textbf{CLIP Prior} &  \textbf{CMD} & \textbf{MCC} & \textbf{RefCOCO} & \textbf{Ref-YT-VOS} \\ \midrule 
            - & - & - & 70.42 {\scriptsize \textcolor{white}{+0.00}} & 59.8 {\scriptsize \textcolor{white}{+0.0}} \\
            \checkmark & - & - & 71.95 {\scriptsize \textcolor{teal}{+1.53}} & 61.5 {\scriptsize \textcolor{teal}{+1.7}} \\ 
            - & \checkmark & - & 73.18 {\scriptsize \textcolor{teal}{+2.76}} & 61.9 {\scriptsize \textcolor{teal}{+2.1}} \\ 
            - & \checkmark &  \checkmark & 75.43 {\scriptsize \textcolor{teal}{+5.01}} & 63.6 {\scriptsize \textcolor{teal}{+3.8}} \\ 
            \rowcolor{gray!15} \checkmark & \checkmark & \checkmark & \textbf{78.16} {\scriptsize \textcolor{teal}{+7.74}} & \textbf{65.4} {\scriptsize \textcolor{teal}{+5.6}} \\
            \bottomrule
            \end{tabular}
            }
            \label{tab:ablation}
\end{minipage}
\begin{minipage}{.01\linewidth}
    \textbf{}
\end{minipage}
\begin{minipage}{.31\linewidth}
    
            \centering
          \caption{\footnotesize{{Ablation on different query initialization methods in CLIP Prior.}}}
            \vspace{-2mm}
          \resizebox{\textwidth}{!}{%
          \setlength{\tabcolsep}{4pt}
            \renewcommand{\arraystretch}{1.2}
          \begin{tabular}{l|l}
            \toprule
            \textbf{Query Initialization method} &  \textbf{RefCOCO} \\ \midrule 
            Only text features~\cite{referformer} &  75.43 {\scriptsize \textcolor{purple}{-2.73}} \\
           \rowcolor{gray!15} \begin{tabular}[c]{@{}c@{}} CLIP Prior with Prompt \\ "A Photo of [Object]"\end{tabular}  &  78.16  \\ 
           \begin{tabular}[c]{@{}c@{}}CLIP Prior mean's performance \\  over  80 ImageNet prompts\end{tabular}
             &  78.25  {\scriptsize \textcolor{teal}{+0.09}} \\ 
             CLIP Prior w.o main object extractor &  74.34   {\scriptsize \textcolor{purple}{-3.82}}  \\
            \bottomrule
            \end{tabular}}
            \label{tab:clipprior_ablation}

\end{minipage}
\begin{minipage}{.01\linewidth}
    \textbf{}
\end{minipage}
\begin{minipage}{.29\linewidth}

        \centering
          \caption{\footnotesize{{Ablation on different bidirectional multimodal fusion modules.}}}
            \vspace{-2mm}
          \resizebox{\linewidth}{!}{%
              \setlength{\tabcolsep}{4pt}
              \renewcommand{\arraystretch}{1.2}
              \begin{tabular}{l|c|l}
                \toprule
               \textbf{Bidirectional fusion} & \textbf{MCC}  &\textbf{RefCOCO} \\ \midrule %
               \rowcolor{gray!15} \textbf{CMD (Ours)} & \cmark &  \textbf{78.16} \\
               ETRIS~\cite{etris} & \cmark &  77.22 {\scriptsize\textcolor{red}{-0.94}}  \\ 
               CoupAlign~\cite{coupalign} & \cmark &  77.01  {\scriptsize \textcolor{red}{-1.15}} \\ 
               CMD & \xmark  &  75.12 {\scriptsize\textcolor{red}{-3.04}} \\
               ETRIS~\cite{etris} & \xmark & 74.12 {\scriptsize\textcolor{red}{-4.04}}  \\ 
               CoupAlign~\cite{coupalign} & \xmark &  73.97  {\scriptsize \textcolor{red}{-4.19}} \\ 
                \bottomrule
                \end{tabular}
            }
            \label{tab:cmd_ablation}
\end{minipage} 
\end{table*}

\subsection{Main Results}

\noindent\textbf{Referring Image Segmentation.}
As illustrated in~\Cref{tab:ref-3datasets}, our method outperforms the state-of-the-art methods by a large margin in all splits of different datasets in the standard setting. Notably, our method surpasses the recent CGFormer and VG-LAW on the validation splits of all three benchmarks, achieving mIoU gains of 1.23\% and 3.11\% on RefCOCO, 1.46\% and 3.31\% on RefCOCO+, and 2.16\% and 4.37\% on G-Ref. The more complex the expressions, the greater the performance gains achieved by VATEX. Compared to LISA~\cite{lisa}, a large pre-trained vision and text encoder, VATEX consistently outperforms it by 3-5\% across all datasets. Furthermore, ~\Cref{tab:precision-analysis} demonstrates the superior performance of VATEX over LAVT, ReLA, and CG-Former on average precision metrics, particularly at the Pr@0.7 and Pr@0.9, illustrating our ability to generate high-quality and complete segmentation masks.

\noindent\textbf{Referring Video Segmentation.} Our model can be extended to video datasets with minor adaptations to handle temporal information. As shown in~\Cref{tab:video-results}, VATEX outperforms current SOTA methods VLT and ReferFormer on the same Video-Swin-B backbone by 1.6 and 3.8 $\mathcal{J}\&\mathcal{F}$ on Ref-Youtube-VOS and Ref-DAVIS17 datasets, respectively. 

\noindent\textbf{Qualitative Analysis.} We provide the visualization of our results in~\Cref{fig:visualize}. Our method can successfully segment objects in complex scenarios, such as the presence of multiple similar objects. For example, in the first column, we can localize the guy who sits in the chair instead of the man standing on the tennis court. In the second sample, VATEX can distinguish the sushi plate among several food dishes that LAVT cannot. In the fourth column, our model can not only identify the correct umbrella belonging to the guy in the light jacket but also segment a part of the shaft of the umbrella that the ground truth does not provide. With the expression "bear child is hugging" in the sixth image, LAVT can only segment the bear's head, and CRIS over-segment to the child and under-segment the bear, but VATEX can output the target bear concisely. 
However, our method fails to segment objects that need to be counted and selected by their order or be described indirectly through another object, as we have not leveraged counting information and object interaction in our model. Another point worth mentioning is the differences in architecture design between VATEX and LAVT. VATEX focuses on instance-based segmentation, while LAVT focuses on pixel-based segmentation. Consequently, VATEX produces smoother and more complete segmentation masks.

\subsection{More Analysis}

\noindent\textbf{Ablation Study.}
We conduct an ablation study on the validation sets of RefCOCO with Swin-B backbone and Ref-Youtube-VOS validation set with Video-Swin-B backbone to examine the impact of each proposed component in our model. The baseline follows the architecture of ReferFormer~\cite{referformer} by using only languages as the initial query (removing CLIP Prior), while only using text-guided vision features and ignoring the vision-aware text features (removing CMD and MCC). As shown in~\Cref{tab:ablation}, the combination of CLIP Prior, CMD, and MCC modules results in the best performance, showcasing a remarkable performance increase of up to 7.74\% in mIoU on RefCOCO and 5.6\% in $\mathcal{J} \& \mathcal{F}$ on Ref-Youtube-VOS. This outcome unequivocally attests to the remarkable effectiveness of our approach and underscores its significant impact. The full ablation study is shown in the supplementary material.  

\vspace{2mm}
\noindent\textbf{Study on different query initialization methods of CLIP Prior.} As described in~\Cref{sec:clip_prior}, our CLIP Prior relies on a template to convert the main noun phrase into a simple sentence suitable for CLIP. As shown in~\Cref{tab:clipprior_ablation}, the baseline follows the query initialization from ~\cite{referformer}, which uses only text features and achieves 75.43\% mIoU. We investigate the effects of using different templates and how these affect the final performance. By using the template ``A Photo of [Object]'', there is a notable improvement of 2.73\% in mIoU. We conducted an additional experiment where we aggregated the text embeddings from 80 ImageNet prompts, which has a very minor performance improvement. This demonstrates that the template ``A Photo of [Object]'' remains a practical choice. We also conduct the experiment that using the whole sentence (instead of the main noun phrase) leads to a significant deterioration in performance at 3.82\%. In this case, CLIP introduces noisy activation on various objects based on their discriminative characteristics within the complex sentence, which is harmful to the model. 

\vspace{2mm}
\noindent\textbf{Different bidirectional multimodal fusion modules}. We ran an ablation study to quantify the performance of CMD with other bidirectional multimodal fusion mechanisms by substituting CMD with alternative modules in ETRIS~\cite{etris} and CoupAlign~\cite{coupalign}. As shown in~\Cref{tab:cmd_ablation}, CMD with MCC performs the best, achieving a mIoU of $78.16\%$, which is higher than both ETRIS~\cite{etris} and CoupAlign~\cite{coupalign} with $0.94\%$ and $1.15\%$, respectively. Disabling MCC results in a notable drop in performance across all settings, with CMD seeing a decrease of $3.04\%$, while ETRIS and CoupAlign experience decreases of $4.04\%$ and $4.19\%$, respectively. This highlights the importance of MCC in improving bidirectional fusion performance, with CMD consistently outperforming alternative methods under the same condition.

\begin{figure}[!t]
    \centering
    \includegraphics[width = 0.99\linewidth]{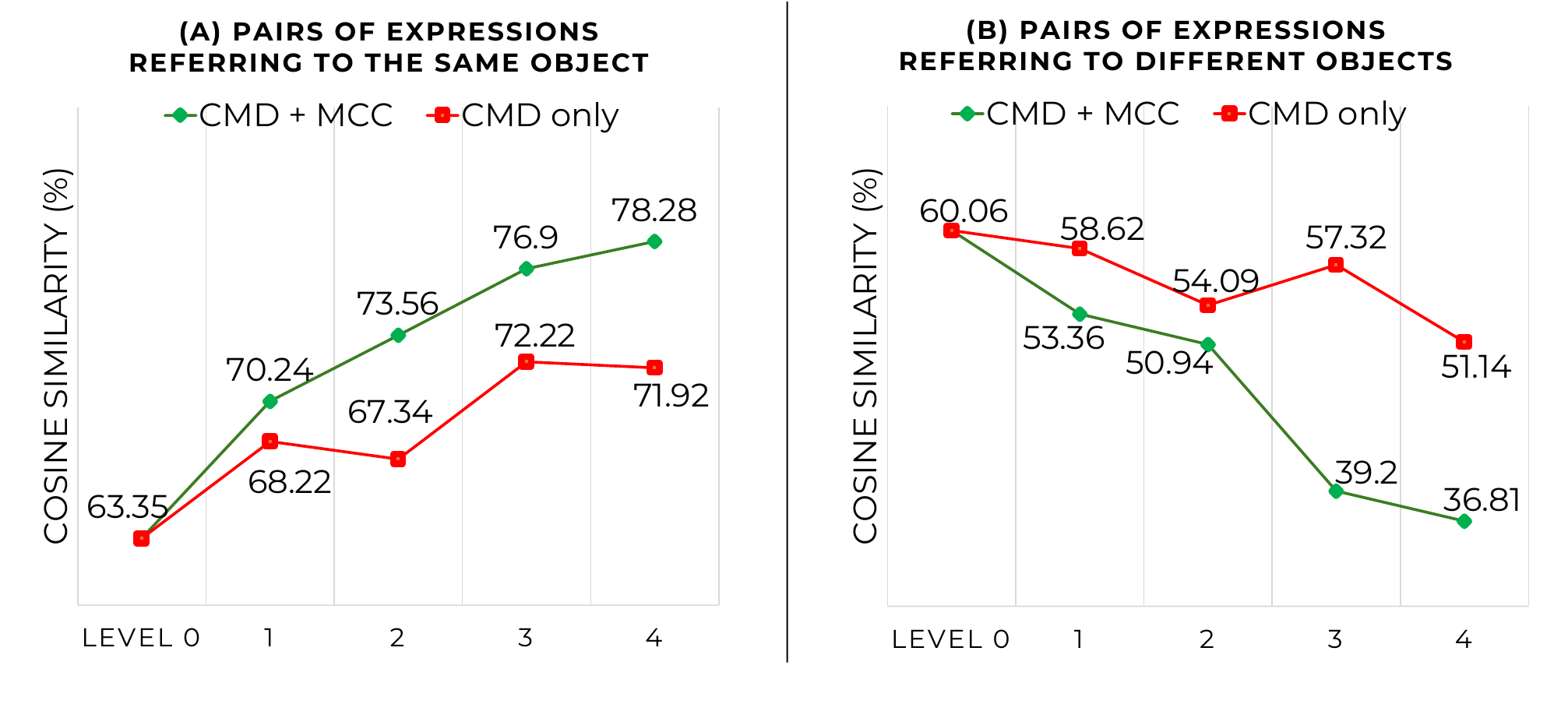}
    \caption{\small Cosine similarities between sentence-level text features at each CMD level. }
    \label{fig:cmd_analysis}
\end{figure}

\vspace{2mm}
\noindent\textbf{Effect of MCC on Vision-aware Text Features.}
To quantify the impact of the MCC on Vision-aware Text Features, we calculate the similarity between sentence-level text features at each layer of CMD, with and without MCC, using the G-Ref dataset. We chose G-Ref because it contains longer, more diverse, and complex context information about the objects expressions, making it ideal for studying the impact of MCC on Vision-aware Text Features. The average similarity results for all pairs of expressions referring to the same or different objects are depicted in~\Cref{fig:cmd_analysis}. Here, level 0 represents the initial features, derived directly from the text encoder, while level 4 indicates the final vision-aware text features of CMD.

In the context of expressions referring to the same object, the initial similarity of text features stands at 63.35\%. By utilizing the MCC, the average similarity gradually increases from level 1 to level 4, reaching 78.28\% at the final vision-aware text feature, while without MCC, the similarity score fluctuates between levels 1 and 3, achieving only $\thicksim$ 8\% performance gain at the end. This illustrates the effectiveness of MCC in guiding the vision-aware text features toward a semantically rich feature space, where the features of two sentences referring to the same object are closer in that feature space.

Regarding different objects, the pairwise similarity of expressions referring to different objects progressively decreases through the four levels of CMD, from 60.06\% to 36.81\%. In contrast, using CMD without MCC results in an unstable feature space. This illustrates the effectiveness of MCC in guiding vision-aware text features toward a more robust feature space, where the features of two sentences referring to different objects can be similar at the beginning but are pushed apart in the final vision-aware text feature.

These findings underscore the pivotal role of MCC in bolstering multimodal comprehension provided by CMD. Specifically, we reveal how the in-context learning signal in MCC effectively improves the representation of vision-aware text features.

\vspace{2mm}
\noindent\textbf{Others.} We reported the Universality of VATEX, the Runtime Analysis of VATEX, and the effect of MCC on object segmentation in Supplementary Material.

\section{Limitations}

Our method is not without limitations. 
Particularly, our method does not exploit positional relationship between different objects as well as the alignment between actions and expressions referring to them (see~\Cref{fig:visualize} the last two columns). Consequently, situations involving counting ("third from left"), indirect descriptions ("kid next to girl in pink pants"), or actions ("a woman walking to the left") might lead to inaccurate predictions. Another line of future work is making RIS work on general scenarios (\eg segment all the red-colored objects, segment all the text in the image) or more fine-grained segmentation (\eg segment the eye of the owl). Dealing with intra-frame object relationships and inter-frame information for video is vital for future work. It is also of great interest to investigate vision-aware text features with the VLMs and to lift this task to the 3D domain.

\section{Conclusion}
\label{sec:Conclusion}

This paper introduces VATEX, a novel framework that examines how vision-aware text features can enhance the performance of RIS by emphasizing on object and context comprehension. First, we propose integrating visual cues into text features during the query initialization process in CLIP Prior for object understanding. First, we propose integrating visual cues into text features during the query initialization process via the CLIP Prior module for object understanding. Second, we exploit the mutual interaction between visual and text modalities through the Contextual Multimodal Decoder (CMD) module and provide an explicit in-context learning signal for the vision-to-language direction using the Meaning Consistency Constraint (MCC). As a result, our proposed method consistently achieves new state-of-the-art results on three benchmark datasets: RefCOCO, RefCOCO+, and G-Ref.

\section*{Acknowledgment} This research was supported by the internal grant from HKUST (R9429). Binh-Son Hua is supported by the Science Foundation Ireland under the SFI Frontiers for the Future Programme (22/FFP-P/11522).


{\small
\bibliographystyle{ieee_fullname}
\bibliography{egbib}
}

\clearpage
\maketitlesupplementary

Our supplementary has 5 sections. \Cref{appendix:implementation} shows additional information about datasets and training procedure. 
\Cref{appendix:VATEX} explains how spaCy is used to extract main noun phrases from sentences and also explains how potential LLMs can be used to create diverse object descriptions to improve dataset annotations.
\Cref{appendix:results} contains the additional experiments on RefCOCO(+/g), Ref-Youtube-VOS and Ref-DAVIS17. This section also illustrates and analyzes the performance of CLIP Prior, CMD and MCC in different situations as well as the runtime and.

\section{Additional Implementation Details}
\label{appendix:implementation}

\subsection{Datasets}

\noindent\textbf{Image datasets.} RefCOCO and RefCOCO+~\cite{refcoco} are two of the largest image datasets used for referring image segmentation. They contain 142,209 and 141,564 language expressions describing objects in images. RefCOCO+ is considered to be more challenging than RefCOCO, as it focuses on purely appearance-based descriptions. G-Ref~\cite{gref}, or RefCOCOg, is another well-known dataset with 85,474 language expressions with more than 26,000 images. The language used in G-Ref is more complex and casual, with longer sentence lengths on average. 

\vspace{2mm}
\noindent\textbf{Video datasets.} Ref-YouTube-VOS~\cite{urvos} and Ref-DAVIS17~\cite{davis} are well-known datasets for referring video object segmentation. Ref-YouTube-VOS contains 3978 video sequences with approximately 15000 referring expressions, while Ref-DAVIS17 consists of 90 high-quality video sequences. These datasets are used to evaluate the performance of algorithms that aim to identify a specific object within a video sequence based on natural language expressions.

\subsection{Metrics} 

In our work, we use mIoU and Precision@X to evaluate our method for image datasets, while $\mathcal{J}\&\mathcal{F}$ are used as evaluation metrics for video datasets. mIoU stands for mean Intersection over Union, which measures the average overlapping between the predicted segmentation masks and the ground truth annotations. Precision@X, on the other hand, measures the success rate of the referring process at a specific IoU threshold, and it focuses on the referring capability of the method.

In addition, region similarity $\mathcal{J}$ and contour accuracy $\mathcal{F}$, and their average $\mathcal{J}\&\mathcal{F}$ are commonly used evaluation metrics for video object segmentation (VOS) datasets. The $\mathcal{J}$ is similar to the IoU score, while the $\mathcal{F}$ score is the boundary similarity measure between the boundary of the prediction and the ground truth. These two metrics together measure the performance of the predicted object mask over the entire video sequence. Higher $\mathcal{J}\&\mathcal{F}$ score indicates better RVOS performance.

Furthermore, to quantify the ability to consistently segment various expressions for the same object and further validate the effectiveness of our proposed Meaning Consistency Constraint, we leverage an Object-centric Intersection over Union (Oc-IoU) score, which calculates the overlap and union area between ground truth and all segmentation predictions of the same object. Specifically, consider the $i$-th object with $K_i$ expressions referring to that object and the corresponding ground truth mask $\text{GT}_i$. Let $P_i^j$ be the model's prediction for the $j$-th expression of the $i$-th object, where $j = \overline{1..\;K_i}.$ The Object-centric IoU can be formulated as follows:
\begin{align}
    \text{Oc-IoU}(\text{GT}_i, \text{P}_i) &= \frac{\text{GT}_i \cap \text{P}_i^1 \cap ... \cap \text{P}_i^{\text{K}_i}}{\text{GT}_i \cup \text{P}_i^1 \cup ... \cup \text{P}_i^{K_i}}, \\ 
    \text{Oc-IoU}_{\text{total}} &= \frac{1}{N}\sum_{i = 1}^{N}{\text{Oc-IoU}(\text{GT}_i, \text{P}_i)},
\end{align}
where $N$ is the total number of objects/instances in the datasets.

\begin{figure*}[!htb]
    \centering
    \includegraphics[width=0.95\linewidth]{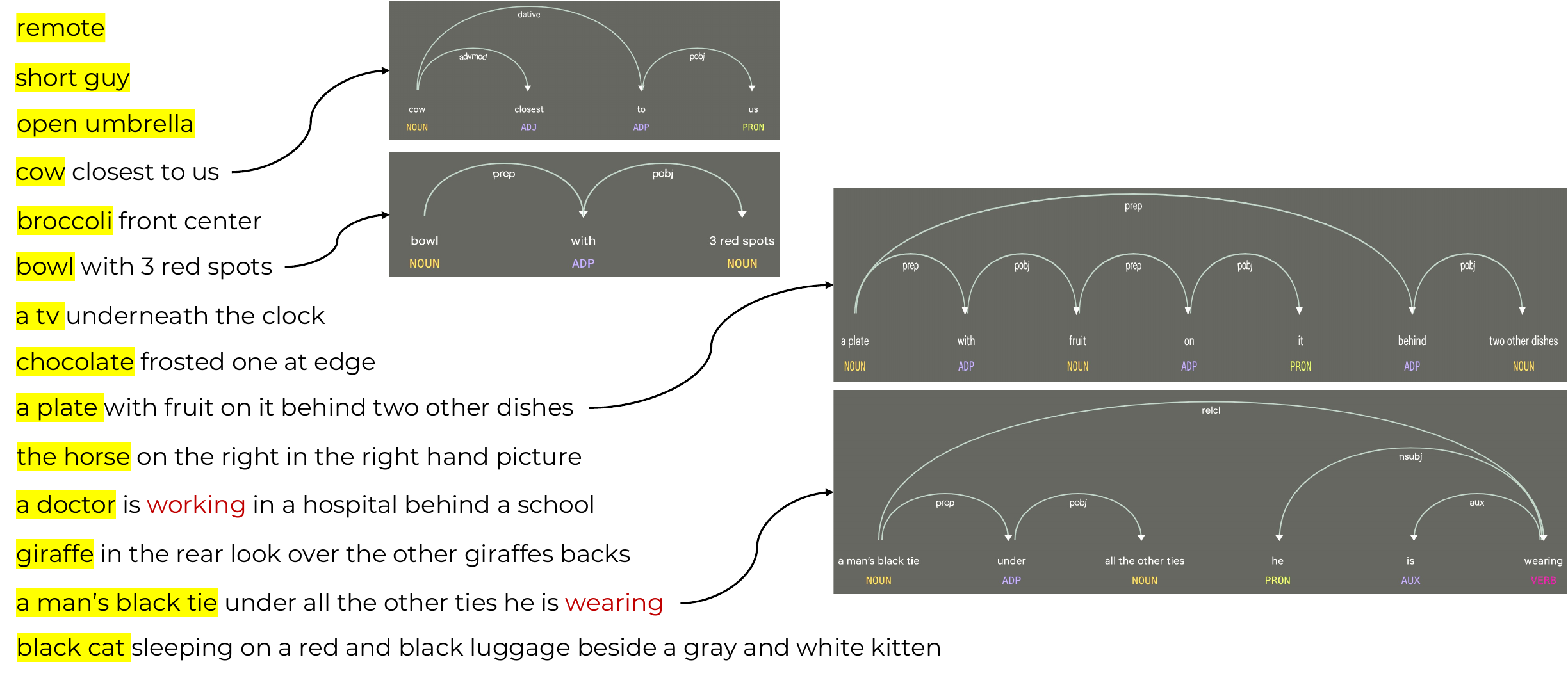}
    \caption{Examples of our main object extractor output. Given the expression, our algorithm will output the \colorbox{yellow}{main noun phrase} in the sentence. Typically, the root word of the sentence is a noun phrase, which we directly output as the main noun phrase. However, if the root word is not a noun phrase (\eg \textcolor{red}{working, wearing} in the image), we instead focus on identifying its child noun. Additionally, we illustrate the dependency parsing tree for some representative sentences on the right.}
    \label{fig:viz_spacy}
\end{figure*}

\subsection{Training Details}

Our model is optimized using AdamW~\cite{adamw} optimizer with the initial learning rate of $10^{-5}$ for the visual encoder and $10^{-4}$ for the rest. Our model comprises a total of nine Masked-Attention Transformer Decoder layers followed~\cite{mask2former}. We set the number of queries to 5~\cite{referformer}. For the setting of training from classification weight from Imagenet on Ref-Youtube-VOS dataset, we train the model for $200,000$ iteration with the learning drop at $140,000$-th iteration. On Ref-DAVIS17\cite{davis}, we directly report the results using the model trained on Ref-YouTube-VOS without fine-tuning. In terms of coefficients in loss function, $\gamma_{cls} = 2$ and $\gamma_{mask} = 5$ are followed from Mask2Former. To maintain balance, we then choose $\gamma_{mcc} = 2$. We want to prioritize the mask loss with the highest weight because the IoU is the primary metric.

\section{Additional Details of VATEX} 
\label{appendix:VATEX}

\subsection{Main Object Extractor} 
\label{appendix:spacy}

We use spaCy \cite{spacy2} to implement our main object extractor, leveraging its optimized, fast, and effective dependency parsing capabilities. First, spaCy extracts the root word of the sentence, also known as the head word, which has no dependency on other words (i.e., it has no parent word in the dependency tree). If this root word is a noun phrase, we directly output it as the main noun phrase of the sentence. If the root word is not a noun (e.g., a verb), we focus on its child noun to ensure it centers on the described object. \Cref{fig:viz_spacy} shows some examples on the datasets and shows the output of our algorithm as well as the dependency parsing tree of some representative cases.

To handle complex sentence structures that lack a directly related noun phrase, we have implemented a rollback mechanism (in L27 of vatex/utils/noun\_phrase.py) that returns the whole sentence, preventing information loss and mitigating potential errors from inaccurate main noun phrase extraction. As shown in \Cref{tab:rollback}, this rollback mechanism helps avoid poorly extracted nouns that could potentially cause incorrect segmentation masks.

\begin{table}[ht]
\centering
\caption{Rollback stats on the validation split of three RIS datasets.}
\label{tab:rollback}
\renewcommand{\arraystretch}{1.15}
\resizebox{0.9\linewidth}{!}{%
    \begin{tabular}{l|ccc}
    \toprule
    \textbf{Dataset}   & \textbf{RefCOCO} & \textbf{RefCOCO+} & \textbf{G-Ref}  \\ \midrule
    Num expressions    & 10,834 & 10,758 & 4,896    \\
    Rollback rate(\%)  & 10.7  & 10.8  & 3.1      \\ 
    mIoU w/o rollback  & 76.23 & 68.45 & 69.01 \\
    mIoU w. rollback   & 78.16 & 70.12 & 69.73  \\
    \bottomrule
    \end{tabular}%
}
\end{table} 

\subsection{Enhancing Expression Diversity in Referring Image Segmentation Datasets through Prompting Techniques}
\label{appendix:llm}

\begin{figure*}[!htb]
    \centering
    \subfloat[Prompting with Mask]{{\includegraphics[width = .49\textwidth]{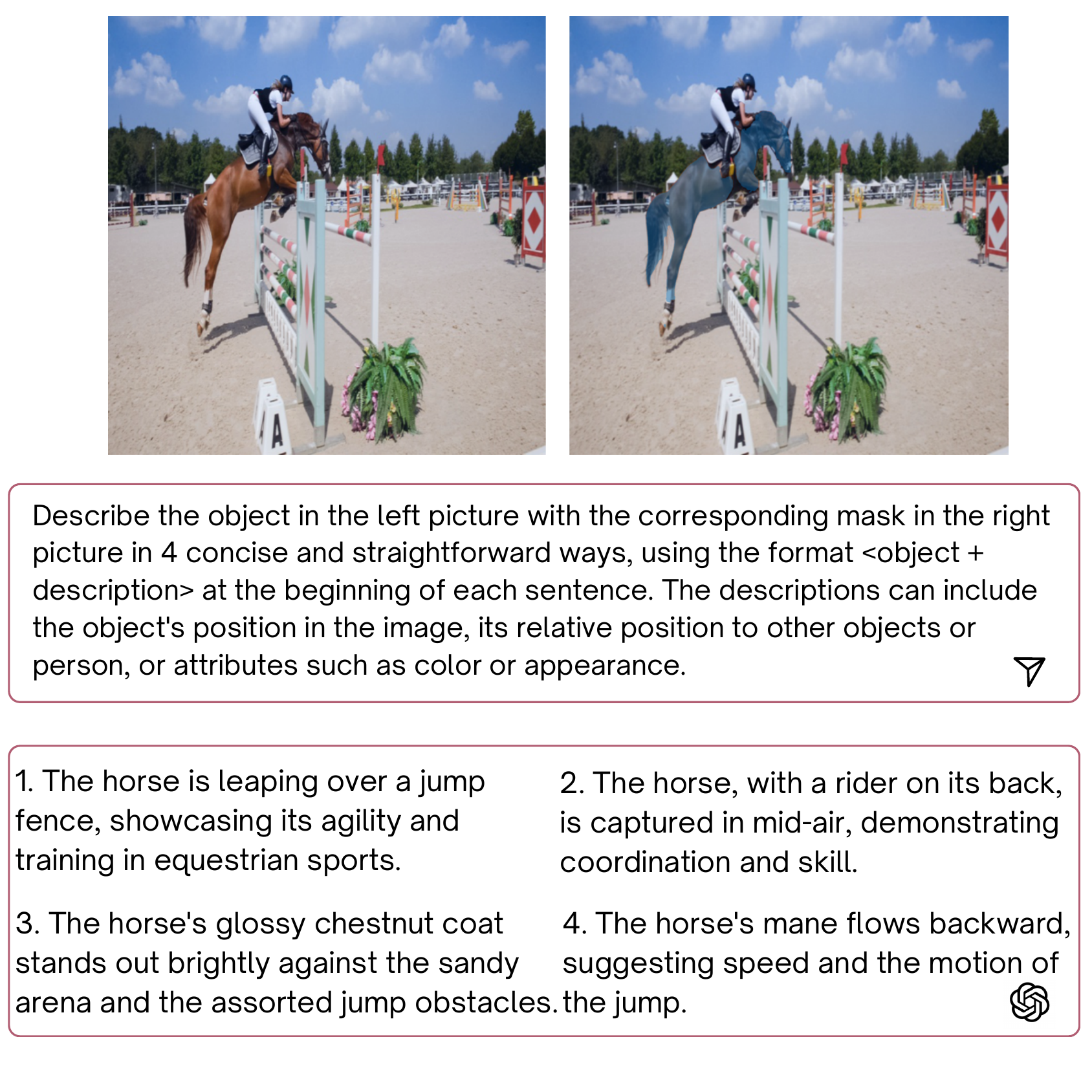}}}%
    \subfloat[Prompting with SoM]{{\includegraphics[width = .49\textwidth]{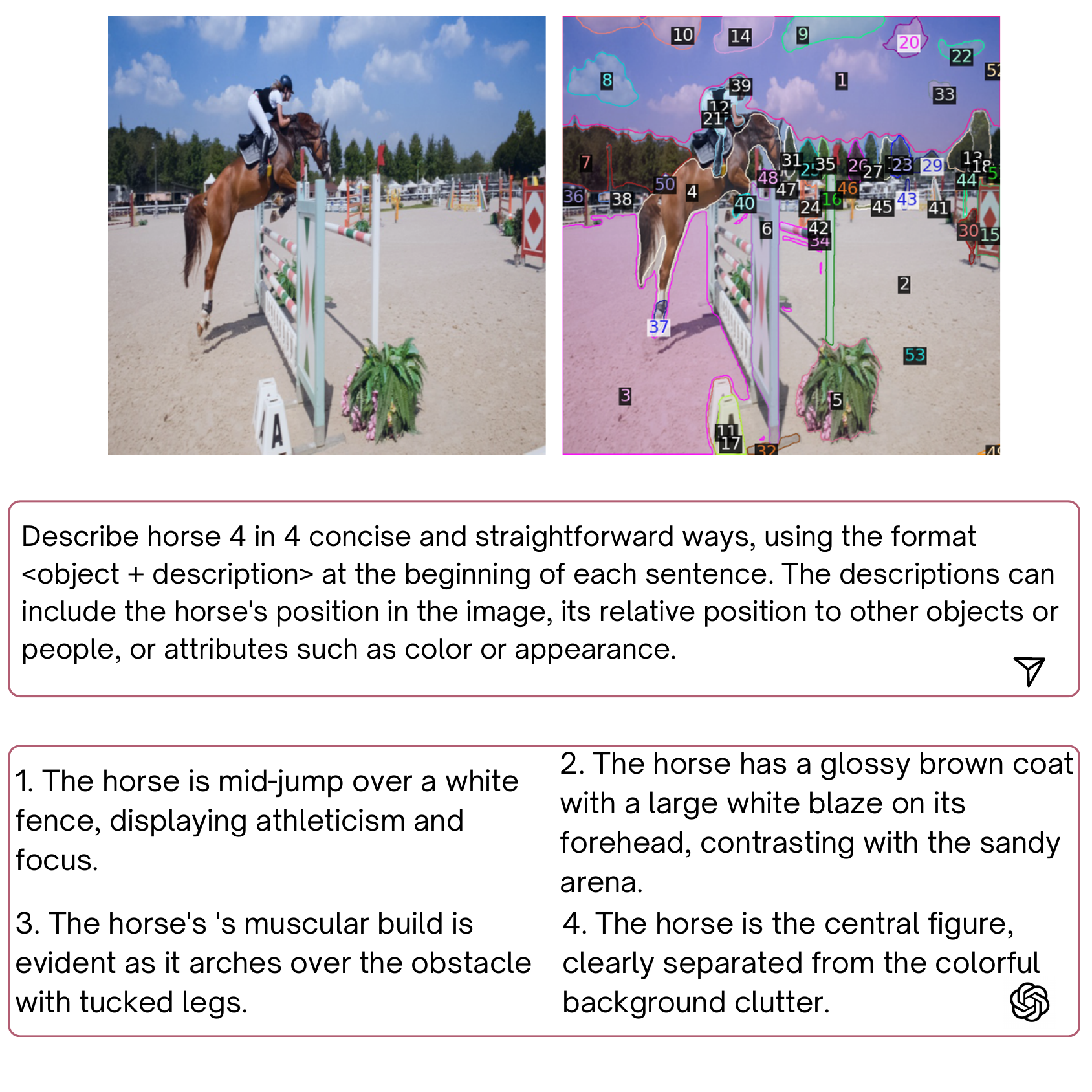}}}%
    \caption{Example of using GPT-4(V) with different prompting techniques to generate object description.}%
    \label{llm}
\end{figure*}

Our method's utilization of diverse referring expressions for each object aligns with established best practices in text-image dataset annotation. This approach is widely accepted and implemented across several benchmark datasets. In scenarios where multiple expressions per object are unavailable, we have the flexibility to employ Large Language Models (LLMs) for enhancing expression diversity. This can be achieved either by augmenting existing expressions or generating new ones based on object masks, a technique successfully employed by datasets like RIS-CQ. Furthermore, we demonstrate a practical application of this approach through a sample that showcases how we can prompt ChatGPT to generate relevant expressions in~\Cref{llm}. This generation is based on factors like an object's position in the image, its relative position to other objects or people, and distinguishing attributes such as color or appearance.

\Cref{llm} showcases two innovative prompting techniques for generating object descriptions. On the left, we demonstrate how combining an original image with its masked version can effectively prompt GPT-4 to generate detailed descriptions. The right side of Figure 10 highlights the application of the SOTA 'Set of Mark' (SoM\footnote{\href{https://github.com/microsoft/SoM}{https://github.com/microsoft/SoM}}~\cite{som}) technique to enhance the capability of GPT-4(V) in acquiring deeper knowledge. SoM involves creating masks for each object in the image using SAM, each distinguished by a unique identifier. This marked image then serves as an input for GPT-4V, enabling it to respond to queries necessitating visual grounding with greater accuracy and relevance.

\section{Additional Results and Analysis}
\label{appendix:results}

\subsection{Universality of VATEX}

VATEX employs CLIP Prior for Object Understanding and Meaning Consistency Constraint for Context Understanding. These two modules can be easily integrated into any DETR-based model (\eg ReLA \cite{rela}) for RIS. We took ReLA as a representative work and reproduced the performance of ReLA on the validation sets of the RefCOCO and G-Ref datasets using mIoU metrics. As illustrated in \Cref{tab:universality}, VATEX seamlessly integrates into current models, achieving significant performance gains of 3.17\% on RefCOCO and 4.05\% on G-Ref. This demonstrates the effectiveness of our approach in utilizing Vision-Aware text features for both object understanding and context understanding.

\begin{table}[t]
\centering
\caption{\small{Universality of VATEX. We conduct experiments to plug-and-play CLIP Prior and MCC in ReLA. \textsuperscript{\textdagger} means we run experiment on their official code to get the mIoU score.}}
\resizebox{0.9\linewidth}{!}{%
\setlength{\tabcolsep}{6pt}
\renewcommand{\arraystretch}{1.15}
\begin{tabular}{l|l|l}
\toprule
{\textbf{Method}} &  {\textbf{RefCOCO}}  &{\textbf{G-Ref}}       \\ \midrule
ReLA\textsuperscript{\textdagger} & 73.16 & 63.64 \\ 
ReLA + CLIP Prior & 74.32   {\scriptsize \textcolor{teal}{+1.16}}  & 65.76    {\scriptsize \textcolor{teal}{+2.12}}  \\ 
ReLA + MCC & 75.46    {\scriptsize \textcolor{teal}{+1.16}}  & 65.12   {\scriptsize \textcolor{teal}{+1.48}}  \\ 
ReLA + CLIP Prior + MCC & 76.33    {\scriptsize \textcolor{teal}{+3.17}}  & 67.69   {\scriptsize \textcolor{teal}{+4.05}} \\ 
\bottomrule
\end{tabular}
}
\label{tab:universality}
\end{table}

\subsection{Additional Comparison on RefCOCO(+/g)} 

\begin{table}[!t]
        \centering
        \caption{\small{{Fair Backbone Comparison between CRIS, JMCELN, LAVT and VATEX.}}}
        \resizebox{0.99\linewidth}{!}{%
        \setlength{\tabcolsep}{4pt}
        \renewcommand{\arraystretch}{1.1}
            \begin{tabular}{l|c|c|ccc}
            \toprule
            \multirow{2}{*}{\textbf{Method}} & \multicolumn{2}{c|}{\textbf{Backbone}} & \multicolumn{3}{c}{\textbf{RefCOCO}}  \\ \cline{2-6} 
                                                &     Visual      &    Textual             & val      & testA   & testB  \\ \midrule
            CRIS~\cite{cris}                & ResNet-101  & CLIP & {70.47} & {73.18}  & 66.10      \\ 
            JMCELN~\cite{JMCELN} & ResNet-101 & CLIP &  74.40 & 77.69 & 70.43 \\
           \rowcolor{gray!15} \textbf{VATEX (Ours)}                   & ResNet-101  & CLIP &\textbf{75.66}	& \textbf{77.88} &	\textbf{72.36}    \\  \midrule
            LAVT~\cite{lavt}                & Swin-B    & BERT & 74.46 &	76.89 &	70.94  \\ 
            LAVT~\cite{lavt}                & Swin-B    & CLIP & 73.15	& 75.24	& 70.02  \\ 
           \rowcolor{gray!15} \textbf{VATEX (Ours)}                   & Swin-B  & CLIP & \textbf{78.16} & \textbf{79.64}  & \textbf{75.64} \\  
            \bottomrule
            \end{tabular}
            }
            \label{tab:backbone-comparison}
\end{table}

\subsubsection{Fair backbone comparison} 

We have benchmarked our model, VATEX, using the ResNet-101 backbone, aligning it with CRIS and JMCELN for a more equitable comparison, as illustrated in ~\Cref{tab:backbone-comparison}. This adaptation demonstrates VATEX's superior performance, achieving a 1.26\% improvement on RefCOCO val and a significant 1.93\% on RefCOCO testB over the current state-of-the-art methods. 

Further, to address comparisons with LAVT, we have experimented with CLIP as the text encoder, adhering to the official repository guidelines. This experiment revealed a performance decline of approximately 1\% when substituting BERT with CLIP as the text encoder. This finding underscores the critical importance of using the CLIP Image Encoder together with the CLIP Text Encoder to maintain model performance. A similar trend was observed with ReferFormer, reinforcing our conclusion. Consequently, when compared to LAVT under the fair conditions in backbone, VATEX shows a substantial improvement, outperforming by 5.01\%, 4.40\%, and 5.62\% on RefCOCO val, testA, and testB, respectively. This data confirms the effectiveness of our approach and the importance of consistent backbone usage for fair and accurate performance assessment.

\subsubsection{External/Multiple Training dataset} 

We compare VATEX with other methods in RIS, which used external training data in ~\Cref{tab:external}. SeqTR~\cite{seqtr}, RefTR~\cite{reftr}, and PolyFormer~\cite{polyformer} enhance their performance on the RefCOCO dataset by incorporating external datasets—Visual Genome with 5.4M descriptions across over 33K categories, Flickr30k-entities with 158K descriptions, and the joint dataset RefCOCO(+/g) with 368K descriptions. Their papers indicate that using such external datasets for pretraining can improve performance by 8-10\%. 

Compared to PolyFormer~\cite{polyformer}, without using external pretraining dataset, VATEX\textsubscript{RefCOCO} demonstrates superior performance over PolyFormer-B, while VATEX\textsubscript{RefCOCO+} and VATEX\textsubscript{G-Ref} achieve comparable results with \cite{polyformer} while using \textbf{42x} and \textbf{69x} smaller datasets respectively, with the exception of the RefCOCO+ test B. The performance's gap on RefCOCO+ Test B, which focuses on non-human objects described purely by their appearance (\eg "the porcelain throne," "part of the bed occupied by a black pamphlet"), could be attributed to the varied object categories covered during the pre-training phase with extensive external datasets. 

On the otherhand, VATEX\textsubscript{joint} adopts a different strategy. By solely utilizing the RefCOCO(+/g) dataset, which is \textbf{16x smaller} than the datasets used by PolyFormer, VATEX\textsubscript{joint} with Swin-B backbone still achieves remarkable results. Specifically, VATEX\textsubscript{joint} outperforms PolyFormer by 4-6\% across all benchmarks, setting a new state-of-the-art result on the RefCOCO dataset. UNINEXT~\cite{uninext} and HIPIE~\cite{hipie}, while achieving strong results, rely on extensive pretraining and data leakage in finetuning (joint training with COCO for segmentation while RefCOCO images and annotations are a subset of COCO train split). In contrast, VATEX achieves competitive performance without relying on such extensive pretraining and removes all potential data leaking in the training phase.

\begin{table*}[t]
    \centering
    \caption{Quantitative results of referring image segmentation on Ref-COCO, Ref-COCO+, G-Ref datasets with other SOTA methods using external training data. VATEX is trained with Swin-B backbone}
    \resizebox{0.95\textwidth}{!}{%
    \setlength{\tabcolsep}{6pt}
    \renewcommand{\arraystretch}{1.2}
    \begin{tabular}{l|c|ccc|ccc|cc}
    \toprule
    \multirow{2}{*}{\textbf{Method}} & \multirow{2}{*}{\textbf{External Datasets}}  & \multicolumn{3}{c|}{\textbf{RefCOCO}} & \multicolumn{3}{c|}{\textbf{RefCOCO+}} & \multicolumn{2}{c}{\textbf{G-Ref}} \\ \cline{3-10} &  
                                                  & val      & testA   & testB   & val      & testA    & testB   & val         & test        \\ \midrule
    SeqTR~\cite{seqtr}$$ & \multirow{3}{*}{\begin{tabular}[c]{@{}l@{}}Visual Genome (5.4M) \& \\ Flickr30k-entities (158K) \& \\ RefCOCO(+/g) (368K)  \end{tabular} } & 71.7	& 73.31	& 69.82	& 63.04	& 66.73	& 58.97 &	64.69	& 65.74 \\ 
    RefTR~\cite{reftr}$$ &   & 74.34	& 76.77 &	70.87	& 66.75 &	70.58&	59.4 &	66.63	& 67.39 \\ 
    PolyFormer-B~\cite{polyformer} & & 75.96 &	77.09 &	73.22 &	{70.65} &	{74.51} &	{64.64} &	69.36 &	69.88 \\
    \midrule 
    UNINEXT-H~\cite{uninext} & Object365 (30M) \&  & 82.2	& -- &	--	& 72.5 &	--&	-- &	74.7	& -- \\ 
    HIPIE~\cite{hipie} & COCO + RefCOCO(+/g) & \textbf{82.6}	& -- &	--	& 73.0 &	--&	-- &	75.3	& -- \\ 
    \midrule 
    \rowcolor{gray!10}\textbf{VATEX}\textsubscript{RefCOCO}     & RefCOCO (142K)       & 78.16 & {79.64}  & {75.64}  & - & -  & -  & - & - \\  
    \rowcolor{gray!10}\textbf{VATEX}\textsubscript{RefCOCO+}     &    RefCOCO+ (141K)     & - & -  & -  & {70.02} & {74.41}  & {62.52}  & -& - \\  
    \rowcolor{gray!10} \textbf{VATEX}\textsubscript{G-Ref}     &    G-Ref (85K)        & - & - & - & - &- & -  & {69.73} & {70.58} \\  
    \midrule 
    \rowcolor{gray!15} \textbf{VATEX}\textsubscript{joint}                  & RefCOCO(+/g) (368K) & 81.53 & \textbf{82.75}  & \textbf{79.66}  & \textbf{74.61} & \textbf{78.75}  & \textbf{68.52}  & \textbf{75.54} & \textbf{76.4} \\  
    \bottomrule
    \end{tabular}
    }
    \label{tab:external}
\end{table*}

\begin{table}[t]
    \centering
    \caption{Quantitative results of referring image segmentation on Ref-COCO, Ref-COCO+, G-Ref validation datasets with SOTA vision foundation models.}
    \resizebox{0.99\linewidth}{!}{%
    \setlength{\tabcolsep}{4pt}
    \renewcommand{\arraystretch}{1.2}
    \begin{tabular}{l|c|c|c}
    \toprule
    {\textbf{Method}} &  {\textbf{RefCOCO}} & {\textbf{RefCOCO+}} &{\textbf{G-Ref}}       \\ \midrule
    Grounded-SAM\cite{groundingdino}\cite{sam} & 67.19	& 58.96	& 59.14	 \\ 
    X-Decoder~\cite{xdecoder} & - & - &  67.5 \\
    SEEM~\cite{seem}  & -	& - &	67.8 \\ 
    \midrule 
    \rowcolor{gray!15} \textbf{VATEX}   & {78.16} & {70.02}  & {69.73} \\
    \midrule 
    \rowcolor{gray!15} \textbf{VATEX}\textsubscript{joint}   & \textbf{81.53} & \textbf{74.61}  & \textbf{75.54} \\  
    \bottomrule
    \end{tabular}
    }
    \label{tab:sam}
\end{table}

\begin{table}[!t]
      \centering
        \caption{Quantitative comparison with the SOTA on Ref-Youtube-VOS.}
        \resizebox{0.99\linewidth}{!}{%
        \setlength{\tabcolsep}{4pt}
		\renewcommand{\arraystretch}{1.2}
            \begin{tabular}{l|c|ccc}
            \toprule
            \multirow{2}{*}{\textbf{Methods}} & \multirow{2}{*}{\textbf{Backbone}} & \multicolumn{3}{c}{\textbf{Ref-Youtube-VOS}}                                           \\ \cline{3-5} 
                         &                    & \multicolumn{1}{c|}{$\mathcal{J}\&\mathcal{F}$} & \multicolumn{1}{c|}{$\mathcal{J}$} &  \multicolumn{1}{c}{$\mathcal{F}$} \\ \midrule 
            \multicolumn{5}{c}{Train with Image segmentation weight from RefCOCO(+/g)}   \\ \midrule  
            ReferFormer~\cite{referformer}  & ResNet-50         & \multicolumn{1}{c|}{55.6}            & \multicolumn{1}{c|}{54.8}       & 58.4       \\ 
            RR-VOS~\cite{rrvos} & ResNet-50 & \multicolumn{1}{c|}{57.3}            & \multicolumn{1}{c|}{56.1}       & 58.4       \\
          \rowcolor{gray!15}  \textbf{VATEX (Ours)} & \textbf{ResNet-50}    & \multicolumn{1}{c|}{\textbf{58.5}}   & \multicolumn{1}{c|}{\textbf{57.1}}  & \textbf{59.9}  \\ \midrule 
            
            ReferFormer~\cite{referformer}  & Swin-L            & \multicolumn{1}{c|}{62.4}            & \multicolumn{1}{c|}{60.8}       & 64.0       \\
          \rowcolor{gray!15}  \textbf{VATEX (Ours)}  & \textbf{Swin-L}    & \multicolumn{1}{c|}{\textbf{64.2}}   & \multicolumn{1}{c|}{\textbf{61.4}}  & \textbf{67.0}  \\ \midrule
            ReferFormer~\cite{referformer}  & Video-Swin-B             & \multicolumn{1}{c|}{62.9}            & \multicolumn{1}{c|}{61.3}       & 64.6       \\
            VLT~\cite{vlt_tpami} & Video-Swin-B           & \multicolumn{1}{c|}{63.8}            & \multicolumn{1}{c|}{61.9}       & 65.6       \\
          \rowcolor{gray!15}  \textbf{VATEX (Ours)}  & \textbf{Video-Swin-B}    & \multicolumn{1}{c|}{\textbf{65.4}}   & \multicolumn{1}{c|}{\textbf{63.3}}  & \textbf{67.5}  \\ \bottomrule
            \end{tabular}%
            }
            \label{tab:ref-youtubevos}
\end{table}

\begin{table}[!t]
    \centering
    \caption{Quantitative comparison with the SOTAs on Ref-DAVIS17 dataset.}
    \resizebox{0.98\linewidth}{!}{%
    \setlength{\tabcolsep}{6pt}
    \renewcommand{\arraystretch}{1.2}
    \begin{tabular}{l|c|ccc}
    \toprule 
    \multirow{2}{*}{\textbf{Methods}} & \multirow{2}{*}{\textbf{Backbone}} & \multicolumn{3}{c}{\textbf{Ref-DAVIS17}}                                           \\ \cline{3-5} 
     & & \multicolumn{1}{c|}{$\mathcal{J}\&\mathcal{F}$} & \multicolumn{1}{c|}{$\mathcal{J}$} & \textbf{$\mathcal{F}$} \\ \midrule
    ReferFormer~\cite{referformer} & ResNet-50           & \multicolumn{1}{c|}{58.5}            & \multicolumn{1}{c|}{55.8}       & 61.3       \\ 
    RR-VOS~\cite{rrvos} & ResNet-50          & \multicolumn{1}{c|}{59.7}            & \multicolumn{1}{c|}{57.2}       & 62.4       \\
    \rowcolor{gray!15}  \textbf{VATEX (Ours)} & ResNet-50 & \multicolumn{1}{c|}{\textbf{61.2}}       & \multicolumn{1}{c|}{\textbf{58.2}}  & \textbf{64.3}  \\ \midrule
    ReferFormer~\cite{referformer} & Video-Swin-B             & \multicolumn{1}{c|}{61.1}            & \multicolumn{1}{c|}{58.1}       & 64.1  \\
    VLT ~\cite{vlt_tpami} & Video-Swin-B             & \multicolumn{1}{c|}{61.6}            & \multicolumn{1}{c|}{58.9}       & 64.3       \\
    \rowcolor{gray!15} \textbf{VATEX (Ours)} & Video-Swin-B    & \multicolumn{1}{c|}{\textbf{65.4}}   & \multicolumn{1}{c|}{\textbf{62.3}}  & \textbf{68.5}  \\ 
    \bottomrule
    \end{tabular}%
    }
    \label{tab:ref-davis}
\end{table}

\subsubsection{Comparison with SOTA foundation models} ~\Cref{tab:sam} illustrates the quantitative performance between VATEX with generalist foundation models: Grounded-SAM\cite{groundingdino}\cite{sam}, SEEM~\cite{seem} and X-Decoder~\cite{xdecoder} in ~\Cref{tab:sam}. For Grounded-SAM, we first use Grounding DINO to extract the bounding box prediction from the text prompt, then we feed that bounding box to SAM to obtain the final segmentation mask. For X-Decoder and SEEM, we directly use the report number on their official github\footnote{\href{https://github.com/UX-Decoder/Segment-Everything-Everywhere-All-At-Once/}{https://github.com/UX-Decoder/Segment-Everything-Everywhere-All-At-Once/}} with Focal-L backbones. While VATEX is trained on much smaller dataset sizes and smaller backbones, VATEX\textsubscript{joint} still significantly outperforms Grounded-SAM with 14.34\%, 15.65\%, and 16.4\% improvements on RefCOCO, RefCOCO+ and G-Ref, respectively. Compared with X-Decoder and SAM, which are trained and finetuned on RefCOCO(+/g) datasets, we also outperform them with approximately 2\% with VATEX and 7.7\% with VATEX\textsubscript{joint}. 

\begin{figure*}[t]
    \centering
    \includegraphics[width = \textwidth]{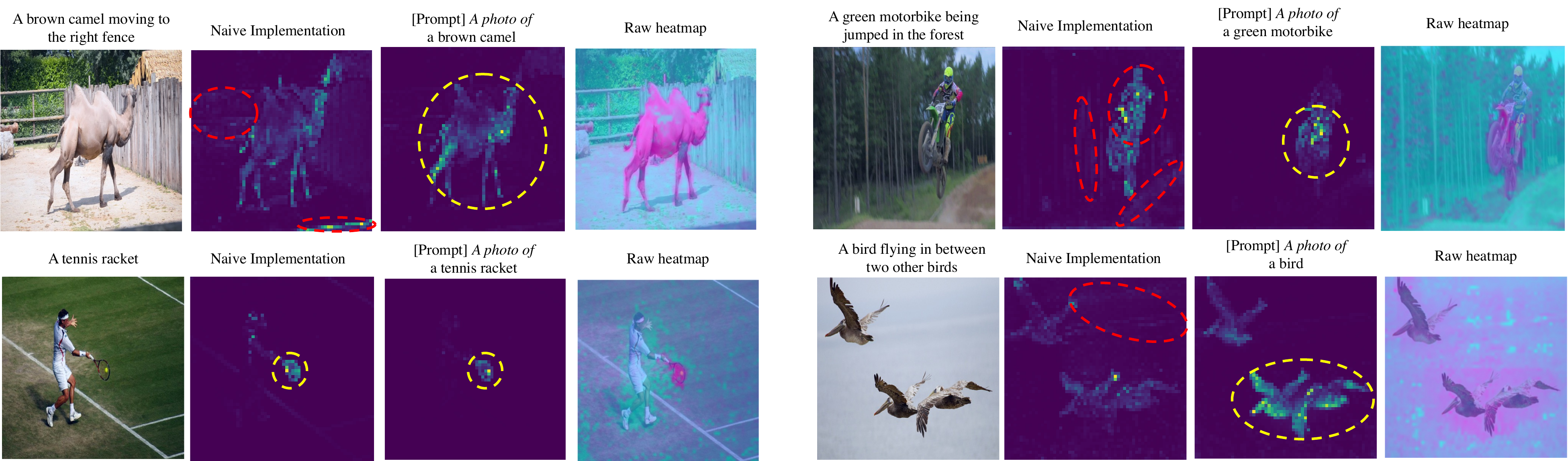}
    \caption{Our heatmap from CLIP Prior. Naive Implementation means feeding the whole sentence through CLIP Model, without the Main Object Extractor. By reducing the complexity of the text expression, it can be seen that the activation on the object of interest becomes more accurate. Best view in zoom.}
    \label{fig:vis_clipprior}
\end{figure*}

\subsection{Experimental results on Ref-YoutubeVOS and Ref-DAVIS17}

The result for Ref-Youtube-VOS dataset is shown in ~\Cref{tab:ref-youtubevos}. 
As can be seen, our method demonstrates superior performance, setting a new state-of-the-art for referring video object segmentation on the Ref-Youtube-VOS dataset with different backbones. In particular, our approach with the spatial-temporal backbone (e.g., Video-Swin~\cite{videoswin}) and pre-trained weights from image dataset achieves the highest $\mathcal{J}\&\mathcal{F}$ score of $65.4\%$ among all other methods on the Ref-Youtube-VOS dataset, including VLT and ReferFormer. 

The results for Ref-DAVIS17 are shown in ~\Cref{tab:ref-davis}. Similarly, our approach achieves competitive performance compared to other state-of-the-art methods in referring video object segmentation. Specifically, with backbones ResNet-50, our proposed model outperforms ReferForme and achieves slightly better results than RRVOS. Moreover, our method achieves the best performance among all methods with the Video-Swin-B backbone with a $\mathcal{J} \& \mathcal{F}$ score of 65.4\%, which is 3.8\% higher than the closest competitor VLT.

\subsection{Heatmap of CLIP Prior} 
\label{appendix:clipprior}

To obtain the heatmap result, from the vector of shape $\left(\frac{H}{16} \times \frac{W}{16} + 1, 1\right)$, we remove "CLS'' token and reshape it into 2D heatmap of $\frac{H}{16} \times \frac{W}{16}$. For visualization purposes, we resize the original image to $960 \times 960$, then pass it through CLIP-Image Encoder, resulting in a high-quality heatmap of size $60 \times 60$. Notably, we only use a default input size of 224×224 during training. Regarding the quality of the heatmap, \Cref{fig:vis_clipprior} demonstrates the comparison between the naive implementation and our prompt-based template. In the 3rd and 7th rows, it is evident that simplifying the sentence and employing prompt templates can aid in distinguishing the target object from the image, resulting in decreased localization errors.

While CLIP Prior excels at localizing objects of interest, it can struggle in complex cases where the expression describes multiple instances within the same category and their relative positions (\eg bottom right of ~\Cref{fig:vis_clipprior}). In these situations, the heatmap may encompass all objects within the category rather than the specific referred instances. However, CLIP Prior's core purpose is to narrow down the relevant region, not pinpoint the exact object. Identifying the precise instance will be handled later in the full-text prompt by the Transformer architecture, which can leverage additional context and relationships.

\begin{figure*}[t]
    \centering
    \includegraphics[width = 0.99\textwidth]{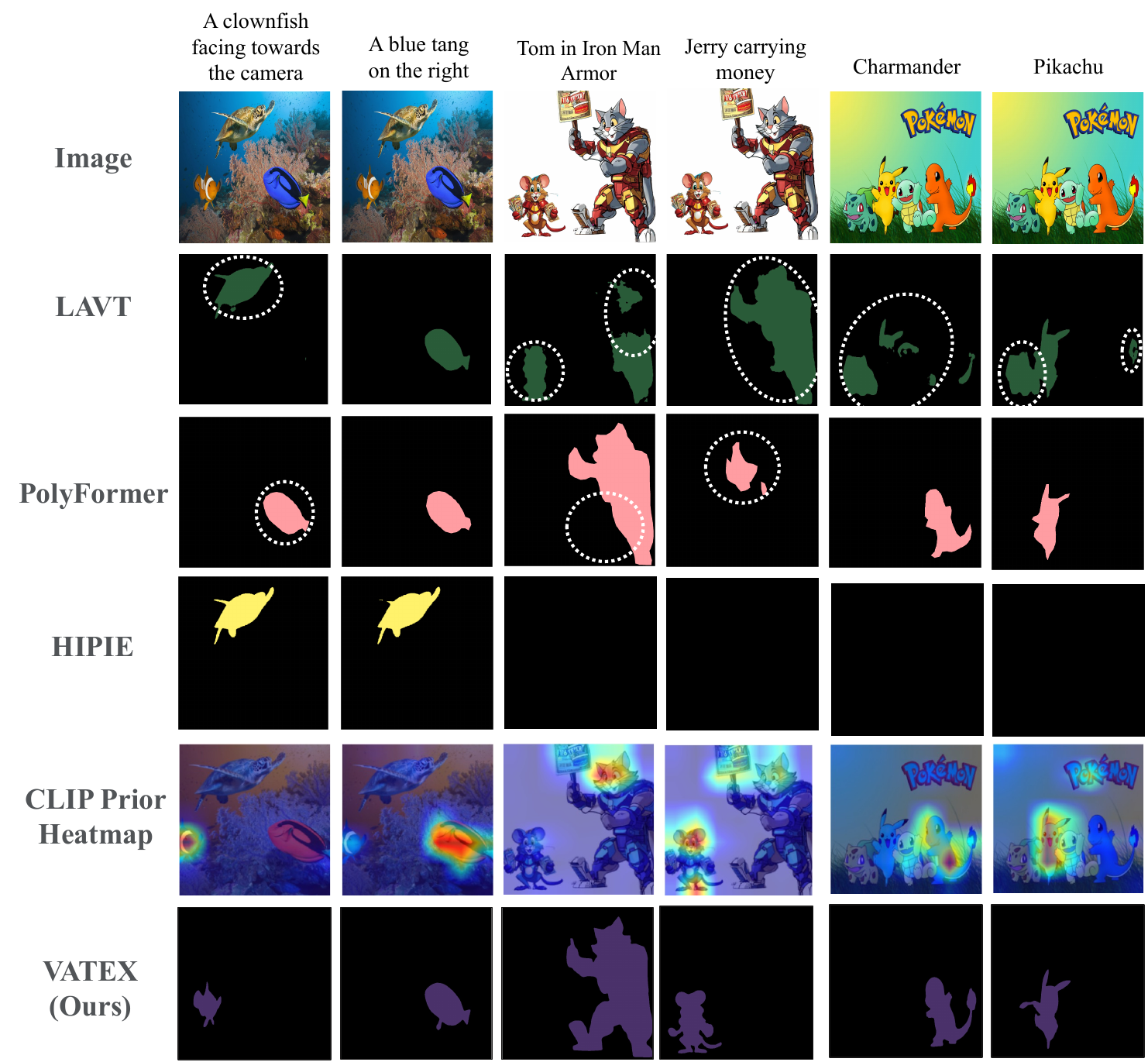}
    \caption{Comparison between VATEX with state-of-the-art methods on challenging out-of-vocabulary cases in referring image segmentation. LAVT's pixel-based approach results in imprecise masks with irrelevant pixel activation. PolyFormer, while creating instance-based masks, struggles with hard cases like "clownfish" or "Jerry" due to limited recognition of unfamiliar objects. HIPIE~\cite{hipie} fails completely due to its constrained pretraining on 365 categories from Objects365. Its high performance on RefCOCO may stem from overfitting and potential data leakage when joint training with COCO. In contrast, VATEX successfully segments correct objects in these difficult vocabulary situations by leveraging the CLIP Prior heatmap. This demonstrates VATEX's superior generalization to unseen objects and complex expressions, highlighting its effectiveness in real-world referring image segmentation tasks. }
    \label{fig:clipprior_openvocab}
    \vspace{-4mm}
\end{figure*}

Moreover, CLIP Prior can also help the model in cases when the referring expression contains out-of-vocabulary objects. By transferring the knowledge from CLIP and embedding the heatmap into the query initialization, the model can obtain a good segmentation mask based on the cues from CLIP Prior. \Cref{fig:clipprior_openvocab} shows how CLIP Prior heatmap can help the model to localize the object in the early phase, thus improving the model's performance. 

\vspace{2mm}
\noindent\textbf{CLIP-based model in RIS.} Adopting CLIP is a good practice taken by several previous methods, including CRIS, CM-MaskSD, and RIS-CLIP. However, to effectively use the aligned embedding from CLIP to obtain good results in referring segmentation is an open question. For example, although using powerful CLIP as the backbone, the SOTA CLIP-based method RIS-CLIP~\cite{risclip} has a comparable performance with the SOTA Non-CLIP model VG-LAW~\cite{vglaw}. To analyze it, we take CRIS~\cite{cris} as a baseline for CLIP-based model. CRIS directly used the well-aligned embedding space between text and vision for RIS. However, the performance of this work is not good compared to others, as there are two concerns with relying solely on CLIP for referring image segmentation tasks:
\begin{itemize}
    \item[1.] Frozen CLIP Model. CLIP model, trained on object-centric images, generates visual features focusing on semantic class meanings rather than instance-based details (see bird example in ~\Cref{fig:vis_clipprior}). This limits the effectiveness of CLIP for instance-level tasks.
    \item[2.] Fine-tuning CLIP Model. Fine-tuning the CLIP model risks overfitting on training samples, thereby diminishing its ability to generalize features to novel classes.
\end{itemize}

We found that learning from a visual backbone pre-trained on ImageNet and only utilizing frozen CLIP as a prior gave better performance on both instance-level segmentation and open-vocabulary segmentation nature of RIS task.

In ~\Cref{tab:clip_comparison}, for a truly fair comparison, we provide our method w/o CLIP, which achieves 75.43, 67.38, and 68.12 mIoU, and we still outperform the SOTA LAVT (74.46, 65.81, and 63.34) and VG-LAW (75.05, 66.61 and 65.36) on RefCOCO(+/g) in the same setting.

\begin{table}[t]
\centering
\caption{Quantitative results of referring image segmentation on Ref-COCO, Ref-COCO+, G-Ref validation datasets on CLIP-based and Non-CLIP model.}
\resizebox{0.9\linewidth}{!}{%
\setlength{\tabcolsep}{4pt}
\renewcommand{\arraystretch}{1.2}
\begin{tabular}{@{}l|cccc@{}}
\toprule
\textbf{Method} &  \textbf{RefCOCO} & \textbf{RefCOCO+} & \textbf{G-Ref} &  \\ \midrule 
\multicolumn{4}{c}{\textbf{CLIP-based Model}}   \\ \midrule 
  CRIS~\cite{cris} & 70.47 & 62.27 & 59.87 \\ 
   CM-MaskSD~\cite{cmmasksd} & 72.18 & 64.47 & 62.67 \\
    RIS-CLIP~\cite{risclip} & 75.68 & 69.16 & 67.62 \\
    \rowcolor{gray!15}{Ours w/ CLIP Prior} &  \textbf{78.16} & \textbf{70.02} & \textbf{69.73} &  \\ 

\midrule
\multicolumn{4}{c}{\textbf{Non-CLIP Model}}   \\ \midrule 
LAVT~\cite{lavt} & 74.46 & 65.81 & 63.34 &  \\
VG-LAW~\cite{vglaw}  & 75.05 & 66.61 & 65.36 \\ 
 \rowcolor{gray!15}{Ours w/o CLIP Prior} & 75.43 & 67.38 & 68.12 & \\ 
\bottomrule
\end{tabular}
}
\label{tab:clip_comparison}
\end{table}

\subsection{Full ablation study}
\label{appendix:ablation}

\Cref{tab:ablation-full} presents an ablation study conducted on the validation set of RefCOCO and Ref-Youtube-VOS, evaluating the mIoU (mean Intersection over Union) and $\mathcal{J} \& \mathcal{F}$, respectively of different model configurations. The study explores the impact of three components: CLIP Prior, CMD (Contextual Multimodal Decoder), and MCC (Meaning Consistency Constraint). 

The first row represents the baseline model with none of the studied components incorporated. The mIoU for this configuration is $70.42\%$ mIoU and 59.8 $\mathcal{J} \& \mathcal{F}$. In rows 2 to 4, the ablation study reveals that incorporating independently the CLIP Prior alone (row 2) and CMD (row 3) both contribute positively to the mIoU on the RefCOCO and $\mathcal{J} \& \mathcal{F}$ on Ref-YoutubeVOS validation set with an improvement of $1.53\%, 2.76\%$ mIoU and $1.7\%, 2.1\%$ $\mathcal{J} \& \mathcal{F}$, whereas the introduction of the Meaning Consistency Constraint (MCC) alone (row 4) leads to a modest increase (only $0.30\%$ mIoU and $0.4$ $\mathcal{J} \& \mathcal{F}$), emphasizing the individual significance of each component in enhancing model performance. Although MCC alone has a modest impact, when combined with the CMD in row 7, there is a notable improvement of $4.7\%$ (mIoU of $75.1$) and $3.3\%$ ($\mathcal{J} \& \mathcal{F}$ of 63.1). This synergy demonstrates that while MCC alone may not perform exceptionally, its collaboration with CMD effectively enhances model performance, aligning with our approach of leveraging enriched text features conditioned by visual information for improved mutual interaction. The final row represents the model with all components (CLIP Prior, CMD, and MCC) combined, achieving the highest mIoU of 78.16 (+7.74) and $\mathcal{J} \& \mathcal{F}$ of 65.4 (+5.6).

\begin{table}[t]
\centering
\caption{Ablation Study on the validation set of RefCOCO (mIoU) and Ref-Youtube-VOS ($\mathcal{J} \& \mathcal{F}$).}
\resizebox{0.99\columnwidth}{!}{
\setlength{\tabcolsep}{4pt}
\renewcommand{\arraystretch}{1.2}
\begin{tabular}{@{}cccc|c|c}
\toprule
 & \textbf{CLIP Prior} &  \textbf{CMD} & \textbf{MCC} & \textbf{RefCOCO} & \textbf{Ref-Youtube-VOS}\\ \midrule 
1 & - & - & - & 70.42 {\scriptsize \textcolor{white}{+0.00}} & 59.8 {\scriptsize \textcolor{white}{+0.0}} \\
2 & \checkmark & - & - & 71.95 {\scriptsize \textcolor{teal}{+1.53}} & 61.5 {\scriptsize \textcolor{teal}{+1.7}} \\ 
3 & - & \checkmark & - & 73.18 {\scriptsize \textcolor{teal}{+2.76}}& 61.9 {\scriptsize \textcolor{teal}{+2.1}} \\ 
4 & - & - & \checkmark & 70.70 {\scriptsize \textcolor{teal}{+0.30}} & 60.2 {\scriptsize \textcolor{teal}{+0.4}}\\ 
5 & \checkmark & \checkmark & - & 75.12 {\scriptsize \textcolor{teal}{+4.72}} & 63.1  {\scriptsize \textcolor{teal}{+3.3}}\\ 
6 & \checkmark & - & \checkmark & 72.14 {\scriptsize \textcolor{teal}{+1.74}} & 61.3 {\scriptsize \textcolor{teal}{+1.5}}\\ 
7 & - & \checkmark &  \checkmark & 75.43 {\scriptsize \textcolor{teal}{+5.01}} & 63.6 {\scriptsize \textcolor{teal}{+3.8}}\\ 
8 & \checkmark & \checkmark & \checkmark & 78.16 {\scriptsize \textcolor{teal}{+7.74}} & 65.4 {\scriptsize \textcolor{teal}{+5.6}} \\
\bottomrule
\end{tabular}
}
\label{tab:ablation-full}
\end{table}

\begin{table}[!t]
        \centering
          \caption{Ablation on the number of queries.}
          \resizebox{0.99\linewidth}{!}{%
              \setlength{\tabcolsep}{4pt}
              \renewcommand{\arraystretch}{1.2}
              \begin{tabular}{c|c|c|c|c|c|c}
                \toprule
                 \textbf{Number of queries} &  \textbf{1} &  \textbf{3} &      \textbf{5}   &   \textbf{10} &   \textbf{20}&  \textbf{50} \\ \midrule %
                 \textbf{RefCOCO} & 77.23   & 77.84   &  78.16  & 78.02 & 78.11 & 77.91 \\
                \bottomrule
                \end{tabular}
            }
            \label{tab:num_queries}
\end{table} 

\Cref{tab:num_queries} presents the impact of varying query numbers on VATEX's performance for the RefCOCO dataset. The results show that while a single query (N=1) achieves a respectable 77.23\% mIoU, increasing the number of queries generally improves performance. The optimal performance is achieved with 5 queries, yielding 78.16\% mIoU, while the performance slightly decreases for query numbers above 5 (78.02\% for 10, 78.11\% for 20, and 77.91\% for 50 queries). The performance pattern is consistent with ReferFormer~\cite{referformer}'s findings.

\subsection{Effect of MCC on Object segmentation mask.} To validate the effectiveness of our proposed MCC module, we propose to use a new Object-centric Intersection over Union (Oc-IoU) score. Unlike mIoU, which averages the overlap and union area for all segmentation predictions within the \textbf{same image}, Oc-IoU measures the overlap and union area between the ground truth and all segmentation predictions for the \textbf{same object} across different expressions, then averages these values across \textit{all objects in the dataset}. This metric provides an evaluation of the consistency and accuracy of segmentation results across various expressions.

~\Cref{tab:mcc} provides the comparisons between our method and the state-of-the-art method LAVT in Oc-IoU on the validation set of three RIS benchmarks. As can be seen, our method outperforms LAVT in all three datasets.  
Comparing the last two rows of~\Cref{tab:mcc}, we can see that the MCC helps the model, especially CMD to enhance mutual information between textual and visual features to further provide more consistent and accurate segmentation. These results underscore the compelling efficacy of our Meaning Consistency Constraint in resolving language ambiguities, thus improving the segmentation performance.

\begin{table}[ht]
    \centering
    \caption{Performance comparison between LAVT and VATEX on Oc-IoU metric.}
    \resizebox{0.9\linewidth}{!}{%
    \setlength{\tabcolsep}{3pt}
    \renewcommand{\arraystretch}{1.15}
          \begin{tabular}{l|cccc}
    \toprule
    \textbf{Method} &  \textbf{RefCOCO} & \textbf{RefCOCO+} & \textbf{G-Ref} &  \\ \midrule 
    LAVT~\cite{lavt} & 62.51 & 50.79 & 56.01 &  \\
    Ours w/o MCC & 66.42 & 54.92 & 59.25 & \\
    \rowcolor{gray!15} Ours & \textbf{68.20} & \textbf{57.38} & \textbf{61.69} &  \\ 
    \bottomrule
    \end{tabular}
    }
    \label{tab:mcc} \\ 
\end{table}

\begin{figure}[!t]
    \centering
    \includegraphics[width=0.99\linewidth]{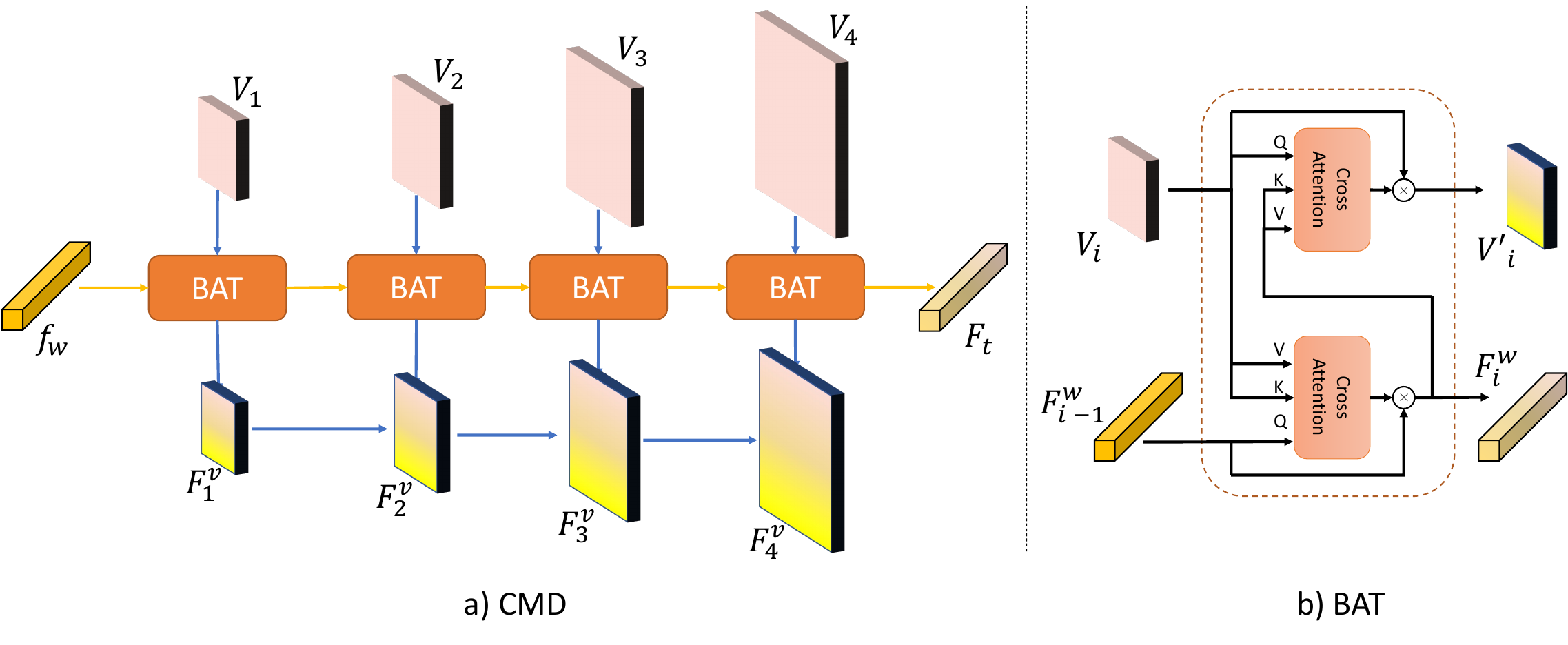}
    \caption{The architecture of Contextual Multimodal Decoder.}
    \label{fig:cmd}
\end{figure}

\subsection{Archiecture Figure of CMD}
\label{appendix:cmd_mcc}

For a robust use of visual and text features in subsequent steps, we propose to fuse visual and text features using a Contextual Multimodal Decoder (CMD), which is designed to produce multi-scale text-guided visual feature maps while enhancing contextual information from the image into word-level text features in a hierarchical design as shown in ~\Cref{fig:cmd}. The process on each level of CMD is achieved by a Bi-directional Attention Transfer(BAT), which incorporates two cross-attention modules.

\subsection{Runtime and Computational Comparison of VATEX}
We report the inference time in FPS and the number of parameters among VATEX, PolyFormer, and LAVT in ~\Cref{tab:extended_comparision}. FPS is measured on an NVIDIA RTX 3090 with a batch size of 1 by taking the average runtime on the entire RefCOCO validation set.

\begin{table}[ht]
\centering
\caption{Comparison in inference time and parameters on the validation set of RefCOCO dataset.}
\label{tab:extended_comparision}
\resizebox{0.99\linewidth}{!}{%
\renewcommand{\arraystretch}{1.15}
\begin{tabular}{l|ccccc}
\toprule
\textbf{Method} &  \textbf{mIoU} & \textbf{FPS} & \textbf{\#params} & \textbf{\#trainable params} \\ \midrule 
LAVT & 74.46 & 13 & 217M & 217M &\\
PolyFormer & 75.96 & 3.5 & 295M & 295M &  \\ 
VATEX\textbf{(Ours)} & 78.16 & 11 & 251M & 165M &\\
\bottomrule
\end{tabular}
}
\end{table} 

\begin{figure*}[!t]
    \centering
    \includegraphics[width=0.95\linewidth]{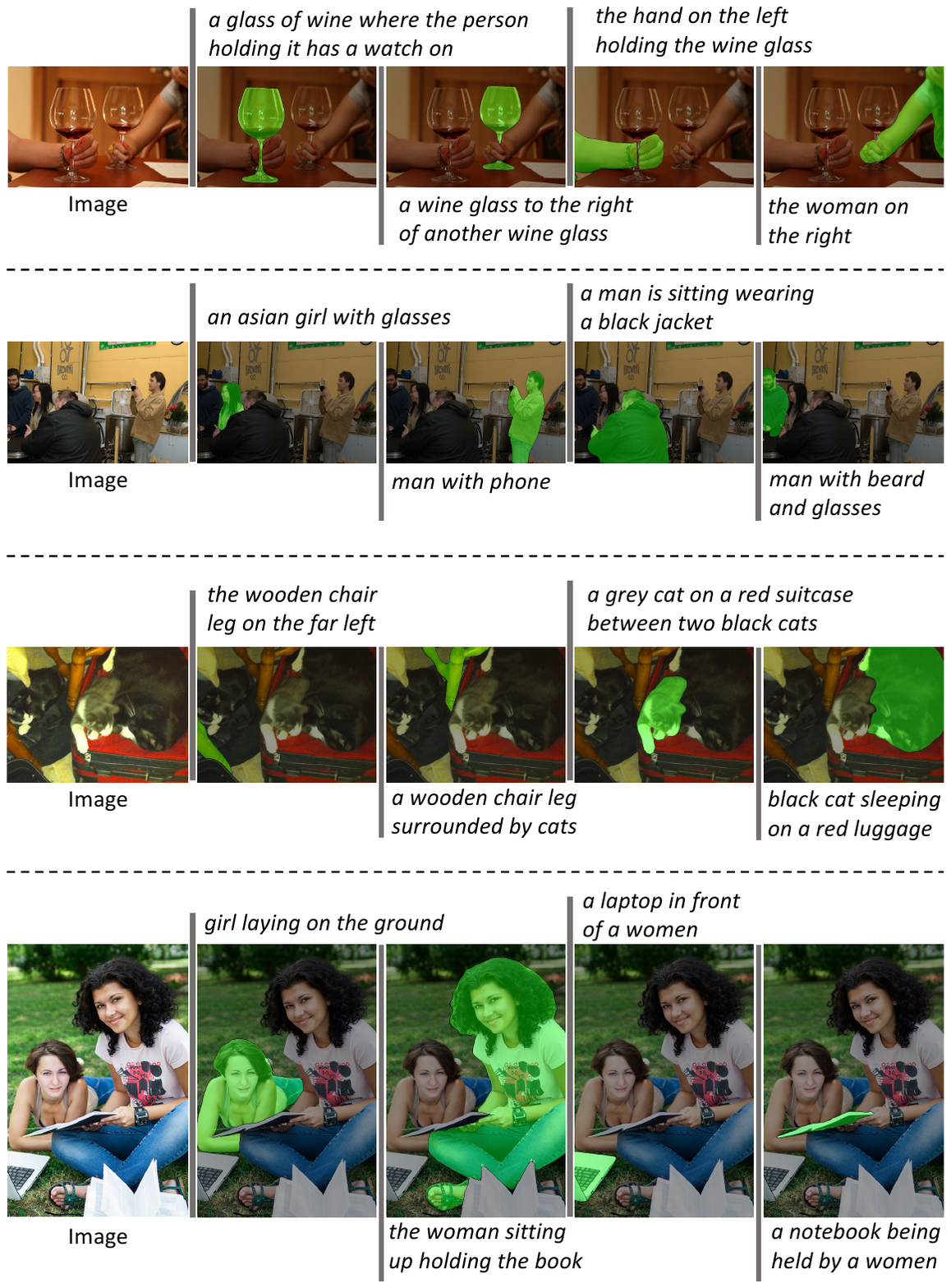}
    \caption{Qualitative results of VATEX according to different language expressions for each image on the validation split of G-Ref.}
    \label{fig:visualization}
\end{figure*}

\begin{figure*}[t]
    \centering
    \includegraphics[width=0.99\linewidth]{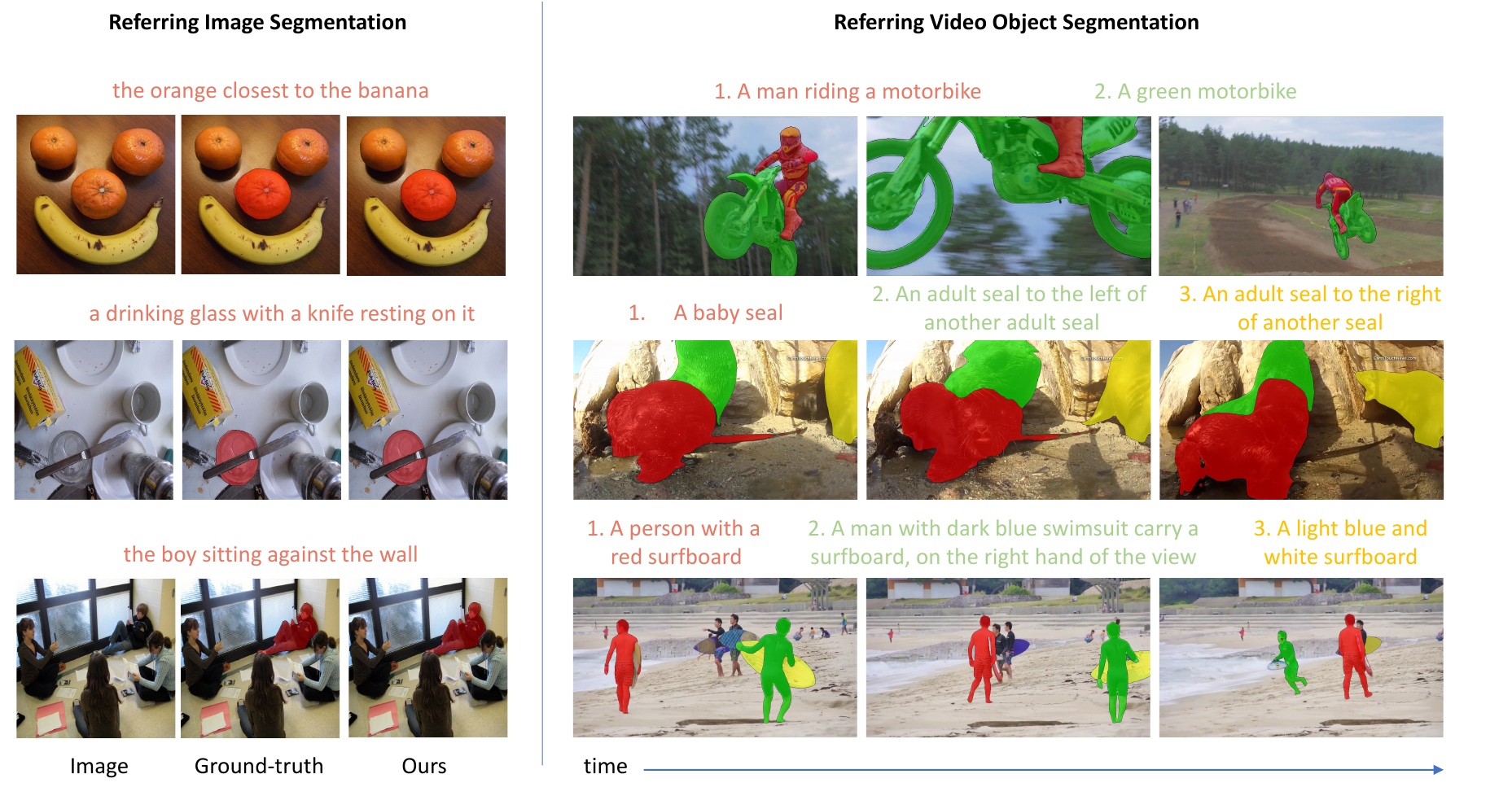}
    \caption{Visualization of  VATEX's results.  VATEX performs well in complex scenarios such as rapidly changing (\textit{motorbike}), and distinguishing from multiple highly similar objects (\textit{people, seal}). The last row of the video results shows a failure case: PDF segments the wrong man in the last column who has similar attributes when the correct one(green) disappears in the video sequences. Best viewed in color.}
    \label{fig:visualization1}
\end{figure*}

\subsection{Additional Visual Results}
\label{appendix:visualization}

In ~\Cref{fig:visualization} and~\Cref{fig:visualization1}, we present additional visualization results for our approach. These results demonstrate that VATEX can successfully segment referred objects in a variety of scenarios, including complex expressions or scenes containing multiple similar objects or rapidly changing shapes. To further illustrate our method's capabilities, we have also created a video demo that compares our approach to ReferFormer on Ref-Youtube-VOS. This video demo is provided as an attachment.

\end{document}